\documentclass[manuscript,screen,nonacm]{acmart}

\AtBeginDocument{%
  \providecommand\BibTeX{{%
    \normalfont B\kern-0.5em{\scshape i\kern-0.25em b}\kern-0.8em\TeX}}}

\setcopyright{acmcopyright}
\copyrightyear{2026}

\usepackage{hyperref}
\usepackage{hyperxmp}
\acmYear{2026}
\acmDOI{XXXXXXX.XXXXXXX}
\usepackage{enumitem}
\setlist{nosep, leftmargin=*}
\usepackage{comment}
\usepackage[utf8]{inputenc}
\usepackage{textgreek}
\usepackage{svg}
\usepackage{graphicx}
\usepackage{booktabs, array,graphicx}
\usepackage{longtable}
\usepackage{subcaption}      
\usepackage{xcolor}
\usepackage{tabularray}

\UseTblrLibrary{booktabs} 
\usepackage{siunitx}
\usepackage{array} 
\usepackage{longtable} 
\usepackage{geometry} 
\usepackage{tikz}
\definecolor{creatorblue}{RGB}{43,92,138}
\definecolor{evaluatororange}{RGB}{217,112,6}
\newcommand{\legendbox}[3][solid]{%
  \fill[#2] (0,0) rectangle (0.12,0.12);
  \ifx\relax#1\relax\else
    \if#1striped
      \foreach \x in {0,0.03,0.06,0.09} {
        \draw[white, line width=0.4pt] (\x,0) -- (\x+0.03,0.12);
      }
    \fi
  \fi
  \node[anchor=west, font=\tiny] at (0.16,0.06) {#3};
}
\usetikzlibrary{positioning, arrows.meta, shapes.geometric}
\usepackage{url}
\usepackage{caption}
\usepackage{graphicx}
\usepackage{array}
\usepackage{ragged2e}
\usepackage{booktabs}
\usepackage{tabularx}

\renewcommand{\arraystretch}{1.1}

\acmPrice{15.00}
\acmISBN{978-1-4503-XXXX-X/18/06}


\begin{document}

\title{What Is the Value of AI-Involved Art? Evidence From a Juried In-Person Exhibition}

\author{Wisdom Obinna}
\affiliation{%
  \institution{Georgetown University}
  \city{Washington}
  \state{D.C.}
  \country{USA}
}
\email{wko7@georgetown.edu}

\author{Hexuan Wang}
\affiliation{%
  \institution{Georgetown University}
  \city{Washington}
    \state{D.C.}
  \country{USA}
}
\email{hw422@georgetown.edu}

\author{Gabriel Lima}
\affiliation{%
  \institution{Max Planck Institute for Security and Privacy}
  \city{Bochum}
  \country{Germany}
}
\email{gabriel.lima@mpi-sp.org}

\author{Cecile Duncan}
\affiliation{%
  \institution{Georgetown University}
  \city{Washington}
    \state{D.C.}
  \country{USA}
}
\email{cld112@georgetown.edu}

\author{Kristelia García}
\affiliation{%
  \institution{Georgetown University}
  \city{Washington}
  \state{D.C.}
  \country{USA}
}
\email{kg870@georgetown.edu}

\author{Nina Grgić-Hlača}
\affiliation{%
  \institution{Max Planck Institute for Software Systems}
  \city{Saarbrücken}
  \country{Germany}
}
\email{nghlaca@mpi-sws.org}

\author{Toni-Lee Sangastiano}
\affiliation{
  \institution{Georgetown University}
  \city{Washington}
  \state{D.C.}
  \country{USA}
}
\email{ts1248@georgetown.edu}

\author{Elissa M. Redmiles}
\affiliation{
  \institution{Georgetown University}
  \city{Washington}
  \state{D.C.}
  \country{USA}
}
\email{elissa.redmiles@georgetown.edu}

\renewcommand{\shortauthors}{Obinna et al.}

\begin{abstract}
What is the value of AI-involved art, and to whom does it belong? We begin to answer these questions using data from an in-person, juried AI art exhibition. Our study involved artists, lawyers, and computer scientists (N=209) as either creators or evaluators of AI-involved artworks. We found that nearly half of evaluators reported willingness to buy an artwork, with pricing correlating with perceived artistic value---creativity and visual appeal---as well as the effort necessary to create it. In contrast, three-quarters of creators reported willingness to sell at prices correlating only with effort rather than artistic value. Both groups viewed prompting users and AI developers as deserving rewards and rights, leaving training data artists behind. Disciplinary training shaped these views: participants with legal training favored distributed reward models. These findings have implications for future research on human-AI creative collaboration and policy debates regarding AI-involved works.


\end{abstract}

\keywords{AI-generated art, copyright, economic value, theory of art, digital media, sweat of the brow}

\maketitle

\section{Introduction}
The use of artificial intelligence (AI) in artistic creation presents a profound transformation to traditional conceptions of creative labor and its value, both culturally and legally~\cite{USCO_WhatIsCopyright, USCO2023_AI_Registration_Guidance, Oksanen2023AIFineArts}. Valuations of art have historically encompassed economic (monetary) and artistic value, often assessed through creativity, visual appeal, and technical quality ~\cite{UNCTAD2024_CreativeEconomyOutlook, Throsby2001EconomicsCulture}. Generative AI (GenAI) disrupts this framework by transforming the nature of artistic labor itself, making creative production faster, cheaper, and less dependent on traditional technical skills. This transformation has raised critical questions about how we value both creative processes and their final products~\cite{messer_co-creating_2024}.

Artists and scholars are divided on what this transformation means for their work and livelihoods. Some artists express growing concern that AI-involved art diminishes the perceived value of creative labor~\cite{Oksanen2023AIFineArts, horton_jr_bias_2023}, viewing current AI training practices as an extraction of original creative value without fair compensation, which threatens artists' livelihoods~\cite{Brewer2024ArtThieves}. In contrast, others view generative AI as a democratizing force that brings cultural creation to the masses and puts creativity in the hands of everyone~\cite{park2025work}. These tensions impact developers who must make consequential choices about the capabilities these systems provide.

Developers and organizations building generative AI face competing demands of advancing the capabilities of generative models while responding to concerns around consent, copyright, attribution, compensation and the effects of AI on creative labor. As a result, the questions about generative AI art extend beyond how artists use these technologies to how the systems themselves are designed, whose interests are prioritized, and who gets the value of the outputs~\cite{Kyi2025}. These competing visions of GenAI's impact present policymakers with difficult choices about whose interests to protect and how~\cite{lovato_foregrounding_2024}.

Policymakers face increasing pressure to develop frameworks that answer this question in a context where the stakes are high and growing. A slew of recent controversies like the U.S. Copyright Office’s partial revocation of protection for \textit{Zarya of the Dawn} after learning its illustrations were AI-generated~\cite{usco2023zarya}; Jason Allen’s first-place win at the Colorado State Fair with an AI-assisted piece~\cite{reuters2024allen}; and Boris Eldagsen’s rejection of the Sony World Photography Award in protest of AI encroachment into creative spaces underscore the need for informed guidance on the right path forward ~\cite{bellaiche_humans_2023}.

Research in human-computer interaction (HCI) has examined the social, ethical, and legal questions raised by AI art from two perspectives. One line of work focuses on developing systems that enable artists to make art with AI without sacrificing their agency in the process~\cite{Verheijden2023, davis2025sketchai, cao2025, almeda2024}. For instance,~\citet{chung2022} designed a tool to allow artists express and refine their creative intentions. Another strand of HCI work has studied how people attribute ownership and copyright for AI outputs~\cite{chung2022, liu2022, lima_public_2025, Epstein2020_WhoGetsCredit}. The results largely show that GenAI systems such as DALL-E, Midjourney, and Stable Diffusion increasingly shift artists' role from directly producing every element of an artwork towards prompting, selecting, iterating, and directing models~\cite{t2i2021dalle, Rombach_2022_CVPR}.

\begin{figure*}[t]
    \centering
    \includegraphics[width=\textwidth]{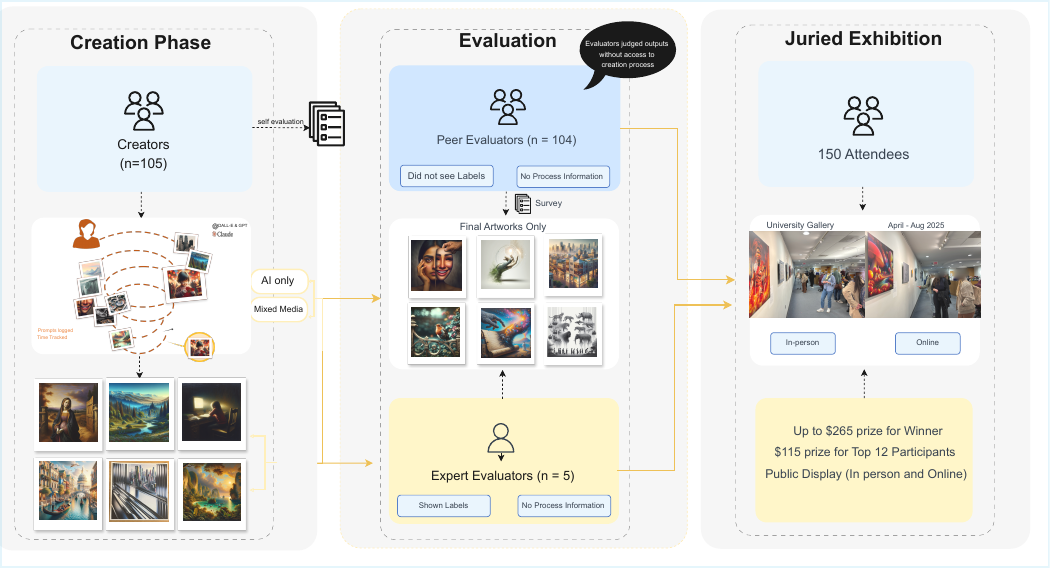}
    \caption{Experimental methodology flow diagram showing two parallel evaluation streams from left to right: (A) Creators ($n=105$) generate artworks using AI-only or mixed-media approaches. (B) Creators produce diverse outputs within the AI-only or mixed-media categories. (C) Creators complete self-evaluations. (D) Evaluators undergo protocol training and assess artworks either with knowledge of the creation process (Experts) or without such knowledge (Peer Evaluators). All images shown in the diagram were selected for exhibition. (E) The 12 selected artworks proceed to the Public and Online Exhibition.}
    \label{fig:experimentOverview}
    \Description{Flow diagram of recruitment, screening, random assignment to conditions, and rating procedure.}

\end{figure*}

Bringing the policy and academic debates together are questions of value and reward~\cite{bautzer2024ubs, USCO2025_AI_Part2_Copyrightability}.
Is art created using generative AI valuable monetarily and/or artistically? If so, who should receive its rewards: the user who prompted the generative AI model, the artists whose art was used to train the model, the AI model itself, the company that owns it, and/or the programmers who built it~\cite{sag2025globalization, Li2025_CopyrightAIChina, USCO2025_AI_Part3_Training}? To answer these questions, we held a juried AI-involved art competition culminating in an in-person exhibition. The competition involved 105 future professional stakeholders who used AI tools in whole or in part to create a submission, as well as 104 peer evaluators and 5 expert evaluators who determined the works exhibited (see Figure~\ref{fig:experimentOverview} Using the data from this exhibition, we address the following research questions:

\begin{itemize}[leftmargin=4em]
\item\textbf{RQ1:} How do future artists, lawyers, and computer scientists value AI-involved works artistically (creativity, visual appeal, technical quality)? How do they value these works monetarily (willingness to buy or sell)?

\item\textbf{RQ2:} How do future artists, lawyers, and computer scientists distribute economic value, authorship, and rights over AI-involved art between stakeholders---e.g., users, model developers, and artists whose art was used to train the model?

\item\textbf{RQ3:} How does training or professional experience in art, computer science, and law influence responses to RQ1 and RQ2?
\end{itemize}

We specifically focused on the perceptions of future professionals because they are poised to shape the solutions to address the creative, technical, and legal questions raised by AI in the future, and because this population enabled an embodied exhibition within a locality in which participants were already invested. Furthermore, this population remains understudied: existing research in HCI has largely examined the general public~\cite{horton_jr_bias_2023, bellaiche_humans_2023, van_hees_human_2025, lima_public_2025}---and sometimes established professionals~\cite{lovato_foregrounding_2024, dacs2024survey}---but little is known about how those currently training to enter relevant fields consider questions of AI-involved art valuation.

By creating a professionally installed juried exhibition in a physical public gallery space, we aim to approximate the conditions under which creative work is typically evaluated, addressing calls in creativity research for assessments that reflect real-world contexts~\cite{Carbon2020_EcologicalArt, Darda2025_MuseumVsDigital}. Evaluator participants' choices thus contributed to a known embodied communal outcome with real stakes, enhancing participants' sense of responsibility. 

Our design deliberately captures dual perspectives of both creators' and evaluators' viewpoints on identical artworks. This responds to calls for creativity research to bridge the traditional separation between studies of creative production and creative evaluation~\cite{glaveanu2014distributed, sawyer2012explaining}. By examining how the same artifacts are valued differently depending on epistemic access to the creative process, we address core questions about how creative value is constructed and attributed in human-AI collaboration.

We find that when considering artistic value \textbf{(RQ1)}, evaluators rated technical quality most favorably, followed by visual appeal and creativity, with their perceptions of the effort a creator put into making the work correlating with their judgments of the work's creativity. Nearly half of evaluators reported hypothetically being willing to buy a piece that they evaluated, at a price that correlates with their perception of the work's artistic value--- its creativity and visual appeal---and the effort invested to create it. Creators felt they put substantial creativity and effort into their submissions, with their perceptions of creativity correlating with both technical quality and observed effort as measured by total prompt length across all iterations. Three-quarters of creators reported hypothetical willingness to sell their work; their asking prices correlated only with the effort they invested rather than the perceived artistic value of their work. 

Regarding attribution of rewards and rights \textbf{(RQ2)}, both creators and evaluators prioritized compensating the prompting user, while training data artists received the least support for rewards and rights. Creators primarily sought self-compensation (63\% would solely compensate themselves). Evaluators adopted somewhat more distributed models: nearly 20\% included training data artists and approximately 38\% included model providers or developers as shareholders of the value or rights associated with a given AI-involved work. 

Disciplinary training shaped these judgments \textbf{(RQ3)}: legal training increased support for compensating training data artists and technical entities while a training in computer science decreased willingness to compensate creators. Similarly, disciplinary training shaped \textbf{(RQ3)} the hypothetical economic value creators assigned their works: training in art was associated with lower hypothesized asking prices.

In the discussion, we consider the design implications for current generative AI applications, and propose that provenance -- through structured process documentation like model cards -- may enable fairer market valuations and provide the kind of process evidence that legal frameworks increasingly require. We also discuss the human authorship requirement for copyright and how it can be understood not only as a philosophical claim about what creativity is or who is deserving but also as a policy choice about how the market should be structured and conclude that centering copyright on human authorship may be the right approach for market-based reasons rather than philosophical ones.

\section{Related Work}
\label{sec:relatedwork}
Here we summarize work on the artistic and economic value of art, human-AI co-creation, and perspectives on artistic value, copyright, and how different stakeholders view these questions. 

\subsection{Artistic Value}
Questions about how to value art are not new. Scholars have long distinguished between intrinsic value based on inherent aesthetic properties of the work itself and extrinsic value based on contextual factors such as provenance, artist reputation, or market positioning~\cite{Throsby2001EconomicsCulture,sep-value-intrinsic-extrinsic-2025}.

\subsubsection{Intrinsic Value} In artistic evaluation, intrinsic value is commonly attributed to the perceived originality, creativity, and expressive or aesthetic qualities of a work, rather than to the amount of effort or labor expended in producing it~\cite{Amabile1982_Competition, Runco2012Standard}. This aesthetic orientation contrasts with copyright law's approach (discussed in \S\ref{subsec:legal}), which protects originality regardless of effort or aesthetic merit. In this study, we adopt the layperson's conception of creativity~\cite{amabile_social_1982, Runco2012Standard}. We use technical quality to refer to domain-specific expertise and craftsmanship~\cite{Amabile2012_Componential, Kozbelt2004}, and visual appeal to refer to aesthetic properties and perceived aesthetic beauty~\cite{Reber2004, Leder2004Model}.

When AI becomes involved in art creation, questions arise about whether these intrinsic qualities remain the primary basis for evaluation. Empirical studies examining AI-generated art suggest they do: \citet{van_hees_human_2025} ran blind comparison experiments, finding participants sometimes preferred AI images over human-made ones, suggesting no clear line between the visual appeal of AI vs. non-AI created works.  Considering AI-created works alone, \citet{messer_co-creating_2024} conducted three vignette-based online experiments where participants read short descriptions of artworks with varying levels of human input disclosed (concept development, curation, refinement). Works described as including greater human involvement were rated as more creative and valuable, with effects extending to reported willingness-to-pay. 

Similarly, \citet{bellaiche_humans_2023} found across labeling experiments that AI-labeled works were systematically devalued relative to human-labeled ones, even when visual quality was held constant---findings similar to those of \citet{horton_jr_bias_2023} in the context of a large-scale survey and platform data. \citet{lima_public_2025} held an online juried AI art exhibition, finding that participants rated their own works as more creative and effortful than others did, and this bias was strongest when monetary rewards were attached. 

These prior works examine how the general public, recruited as crowdworkers, perceive generative AI outputs in online and decontextualized settings. Instead, we bridge Messer’s connection of artistic and economic value with Lima et al.'s method to study how future professionals in relevant disciplines value and attribute value, authorship, and rights in a realistic setting~\cite{messer_co-creating_2024, lima_public_2025}. In so doing, we offer greater ecological validity than prior online surveys. Appendix~\ref{tab:literature} offers a table further contextualizing our work relative to prior work on AI-generated art.

\subsubsection{Extrinsic Value}
Extrinsic value refers to the economic or financial value of an artwork derived from external factors, such as sale price and valuation~\cite{sep-value-intrinsic-extrinsic-2025}. Studying art made with traditional media, \citet{Newman2012} conducted five survey experiments where participants evaluated original vs. perfect duplicates and consistently found respondents are willing to pay more for something they believe is authentic (genuine, original, verified) even when the duplicate was visually identical and independent of the effort involved. \citet{lee_social_2024} combined sales data with visual feature analysis and social signals (gallery representation, critical reviews, museum exhibitions), showing that social signals were stronger predictors of final prices than artistic features. Thus, external cues can sometimes outweigh the artwork’s inherent properties in determining value. When applied to generative AI, similar signals such as provenance and process transparency are also important: \citet{Li2024_InArtWeTrust} analyzed blockchain-based art transactions and showed two visually identical AI artworks could sell for very different prices if one was accompanied by a clear story of how it was made, or curated by a recognized artist.

\subsection{Human-AI Co-Creation and Perspectives of Value}

HCI research has explored how to design computational systems that support human creative agency. 
For instance, \textit{Drawing Apprentice}, which was designed to draw concurrently with a human partner. 
Similarly, 
\textit{DuetDraw}, which allows users and AI agents to negotiate who leads the drawing process, and 
\textit{Cococo}, which added interactive controls that promote effective co-creative experiences between novice users and AI.

With the rise of Text-to-Image models, this collaboration increasingly shifted toward prompting, with some studies already characterizing prompting as a skill that can be developed by artists and non-artists alike~\cite{liu2022,Oppenlaender2025}. 
A study has found that artists already incorporate AI and prompting in distinct ways, particularly as a resource for ideation and as a means of production~\cite{rajcic2024}.

These systems raise an evaluative problem: when an artifact is produced from contributions by both a person and a machine, what exactly should be evaluated and whose artistic value does the resulting output reflect? Reviewing computational co-creative systems,~\citet{karimi2018evaluating} found that existing evaluations have focused predominantly on the user experience of collaboration rather than on the creativity of the jointly produced artifact, proposing a framework distinguishing who evaluates, what is evaluated, and when evaluation occurs. ~\citet{Epstein2020_WhoGetsCredit} showed that credit allocated to an AI system---instead of the human artist or developer---depends on how anthropomorphic the system is perceived to be and that these allocations can be manipulated by introducing the system as a tool. Similarly~\citet{draxler2024} identified a related gap between private and public attribution: users of AI writing tools rarely report a strong sense of ownership over AI-generated text, yet may still decline to disclose the AI's role when presenting the work to others. These questions on attribution become more consequential when they intersect with rights over the creative work.  ~\cite{lima_public_2025} showed that people make attribution judgments about human and AI contributions when assigning authorship and ownership. Legally speaking, however, this attribution is regulated through copyright law, which determines whose creative contribution qualifies for authorship and, consequently, copyright protection.

\subsection{Legal Protection of Value Through Copyright and Ownership}
\label{subsec:legal}

Copyright law, particularly in the United States, protects artistic works by linking authorship and ownership to an identifiable human contribution. In traditional copyright doctrine for human-created works, the US Supreme Court in \textit{Feist v. Rural Telephone Service} held that originality requires minimal creativity rather than mere labor~\cite{feist1991}, establishing that copyright rewards creative expression, not industrious effort. When AI is involved in the creative process, protection depends on demonstrating sufficient human contribution. U.S. Copyright Office registration data confirm that only human-authored elements in AI-assisted works qualify for protection, with applicants required to disclose and disclaim any portions attributable to generative AI~\cite{USCO2023_AI_Registration_Guidance}. International approaches vary: in the UK, statutory provisions extend authorship to the person making the ``arrangements necessary'' for computer-generated works~\cite{uk_copyright_2024}, while in China, case-level analyses show courts granting protection to AI-generated works when they find intellectual creation in prompt design, selection, or control~\cite{wen2024}.

U.S. Copyright Office practice reinforces this creativity-over-effort standard: recent reports and administrative decisions reveal some refusals of fully machine-generated works, with acceptances only when applicants can demonstrate meaningful human editing or curation~\cite{USCO2025_AI_Part2_Copyrightability, usco2023zarya, USCO2025_AI_Part3_Training}. At the international level, WIPO’s 2023 Global Intellectual Property Perception Survey documents substantial variation in public understanding and attitudes toward intellectual property rights, underscoring that perspectives on copyright and innovation differ worldwide. This variation reinforces the role of institutional and societal practices in shaping how value-adding human actions are recognized within policy frameworks~\cite{wipo2023_perception}.

\subsection{Stakeholder Perspectives on Value and Control}

Debates about how to value and protect AI-generated art involve multiple stakeholder groups with distinct interests and perspectives. Among artists, both practicing creators and those whose works trained these models report that while GenAI improves efficiency and ideation, it erodes artistic autonomy and creates competitive adoption pressures~\cite{shi2023_understanding}. Legal scholars, in turn, evaluate how copyright doctrines should apply: Ginsburg~\cite{Ginsburg2003_AuthorshipConcept} argues human authorship forms the necessary basis for value and rights allocation, a position reinforced by the U.S. Copyright Office, which affirms that identifiable human contribution is required for legal protection, excluding AI-only works~\cite{USCO2025_AI_Part2_Copyrightability}. In contrast, technologists, including the companies, developers, and engineers deploying these models frame value through technical benchmarks that compare model performance to human capabilities. Examples include Imagen's photorealism achievements~\cite{saharia_imagen_2022}, GPT-4's human-parity performance~\cite{openai2024gpt4}, and the Human Preference Score v2 benchmark that aligns aesthetic quality assessments with human evaluators across 800,000 judgments~\cite{wu2023_hpsv2}. While these findings suggest differing conceptions of value across disciplines, it remains unclear how training in arts, law, and computer science shapes comparative judgments of artistic and economic value in realistic settings (RQ1/RQ3) and how this value is distributed across contributing parties (RQ2).

\section{Methods}

We conducted a study in the form of a real, in-person juried AI art exhibition through which we collected participants' perceptions of GenAI artworks. Participants took on two roles: Creators ($n$=105), who generated the artworks and submitted them for exhibition, and evaluators ($n$=104), who judged these submissions. We examined how both groups valued AI art through multiple dimensions: economic value (willingness to accept and willingness-to-pay), artistic value (creativity, visual appeal, technical quality), and effort (perceived and actual creative labor) as well as how they attributed rewards and rights to different stakeholders in the AI art pipeline (prompting users, AI developers, training data artists). These factors directly address our research questions on valuation (RQ1), attribution (RQ2), and the role of disciplinary training in shaping these judgments (RQ3). The study was approved by our institution's Ethics Review Board.

\label{app:interfaces}
\begin{figure*}[t]
    \centering
    \begin{subfigure}[b]{0.48\textwidth}
        \centering
        \includegraphics[width=\textwidth]{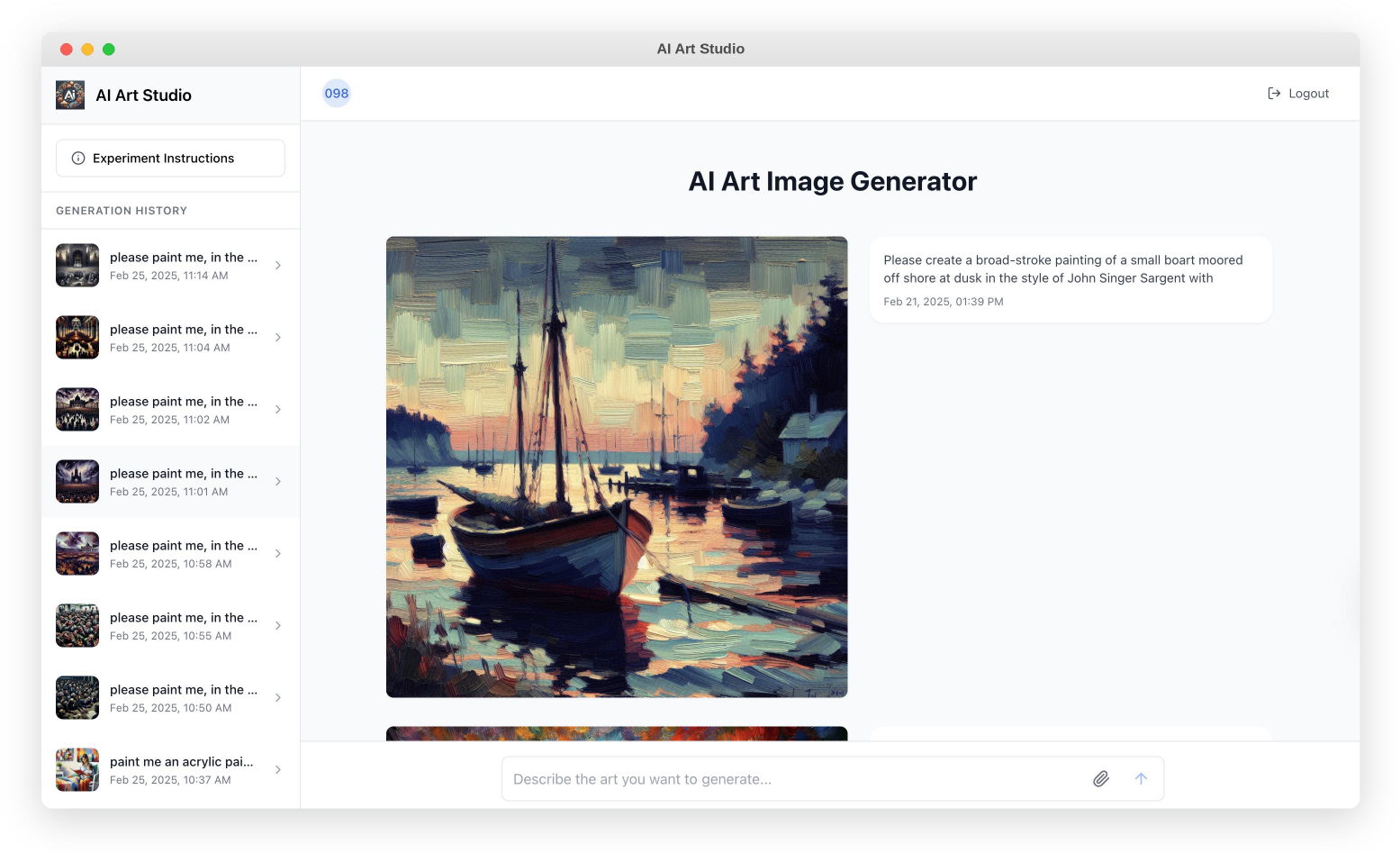}
        \caption{Creator application interface}
        \label{fig:creator_interface}
    \end{subfigure}
    \hfill
    \begin{subfigure}[b]{0.49\textwidth}
        \centering
        \includegraphics[width=\textwidth]{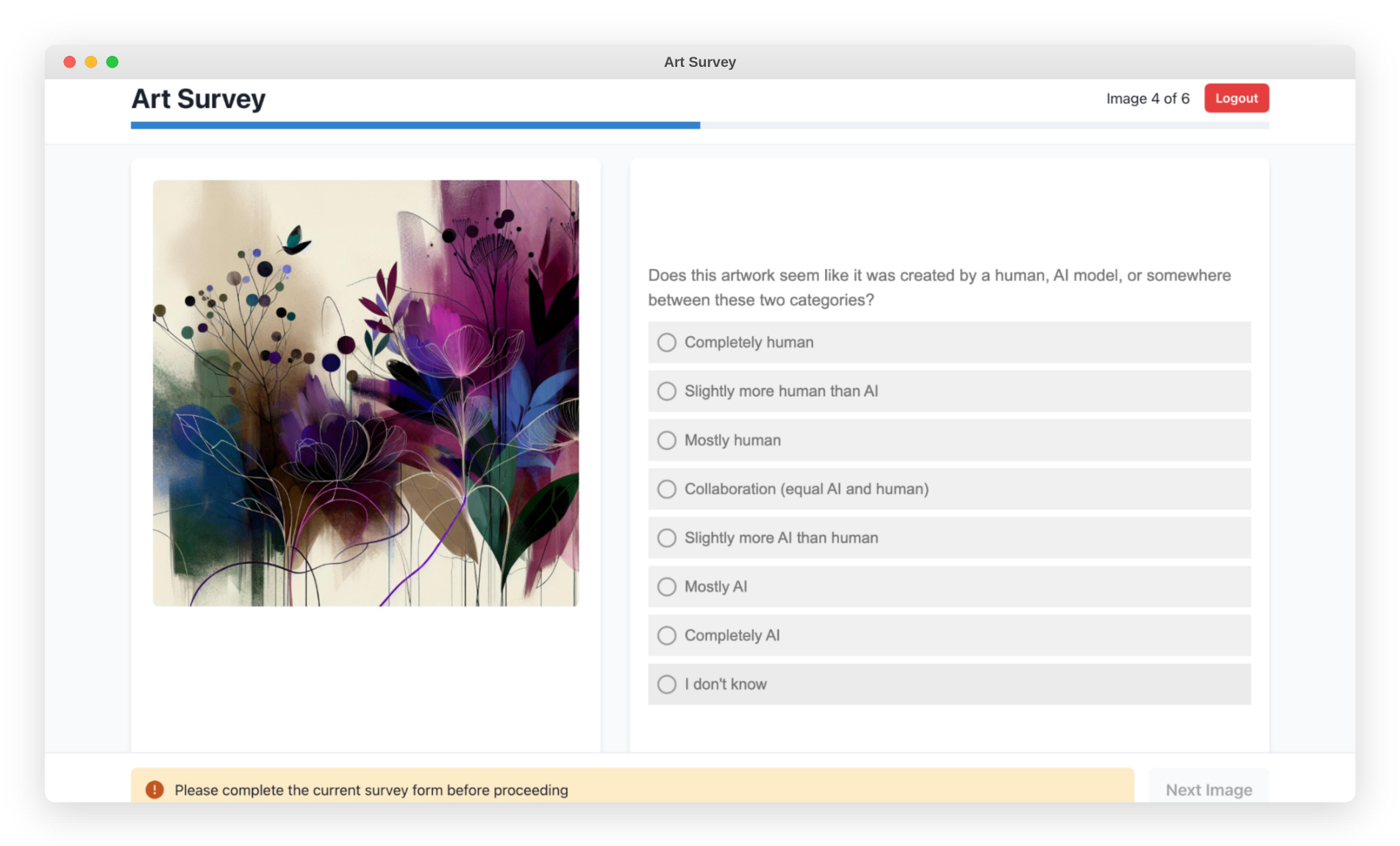}
        \caption{Evaluator survey interface}
        \label{fig:evaluator_interface}
    \end{subfigure}
    \caption{Study interfaces. Left: creator application used to generate artworks via text prompts; the interface automatically logged time, prompt iterations, and prompt text. Right: evaluator survey showing the image-only judgment screen (no origin labels or process information), where participants rated creativity, effort, technical quality, and visual appeal, and then allocated rights and value.}
    \label{fig:interface}
    \Description{Study interfaces}
\end{figure*}

\subsection{Creator Protocol}
\begin{figure*}[t]
    \centering
    \includegraphics[width=\textwidth]{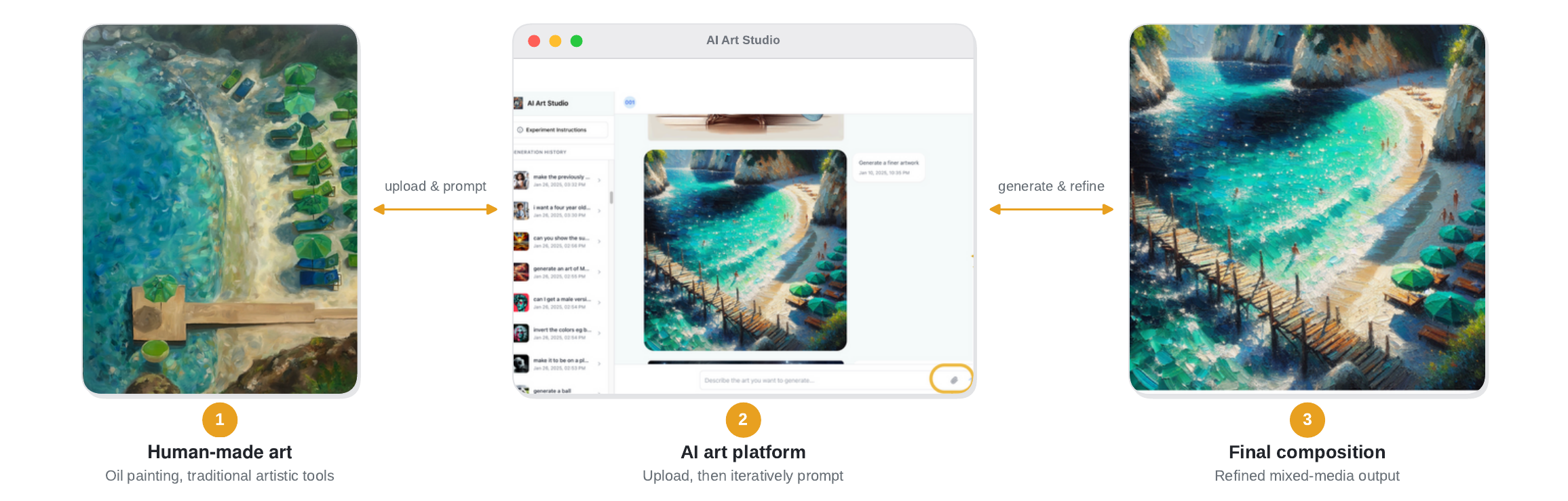}
    \caption{Mixed media category workflow. Creators could choose to begin with traditional handcrafted artwork (left), upload it to the AI art generation platform (center), and iteratively prompt the system to produce their final composition (right). Creators could also choose to begin with the AI art generation platform before continuing with traditional artistic tools.}
    \label{fig:mixed_media}
    \Description{Mixed media workflow}
\end{figure*}

Creator participants could sign up for one of two exhibition categories: 
\begin{enumerate}[label=\alph*)]
\item \textit{AI-only}: Must generate artwork using only Text-to-Image models (specifically our system, see $\S$\ref{subsubsec:platform}) without incorporating traditional artistic techniques or media.
\item \textit{Mixed Media}: Must combine traditional artistic methods, including digital editing, physical media, or other creative techniques with AI (e.g., to enhance or modify their hand-created images, combine AI-generated elements with traditional media to enhance, or post-process AI-generated base images) (see Figure~\ref{fig:mixed_media}).
\end{enumerate}

Creators were told that: (1) the exhibition was juried, meaning submissions would be evaluated by a panel of peer jurors and expert judges and that only the top 12 selected works would be displayed in the exhibition, following the procedures in \S\ref{subsec:exhibition}; (2) in addition to the \$15 for participation, the 12 selected artworks would receive an additional \$100 reward, and the top artwork in each category (AI-only vs. Mixed Media would receive an additional \$150 (total category winner reward: \$265); and (3) the 12 selected works would be showcased at the University Community Gallery exhibition (the opening reception was ultimately attended by 150 people) from April to August 2025, as well as online(see figure~\ref{fig:top-images}).

\subsubsection{Creation Platform} 
\label{subsubsec:platform} 
Creators ($n$ = 105) were required to use a custom web-based system we developed by linking Open AI's DALL-E, and GPT and Claude (see creation interface in~\ref{fig:interface}). The system provided a tutorial on prompt writing, memory features, and category-specific rules. The platform automatically logged prompts, outputs, and session time. 105 creators submitted images created using our interface; submissions that could not be traced back to the online interface were excluded. Following submission, creators were surveyed on their perceptions of the creativity, effort, and technical skill they invested in their submission. Through open-ended responses, they also shared their artistic process, inspiration, and challenges. Creators also made judgments about rights control (who should have decision-making authority over their work) and value distribution (how any monetary reward they received should be hypothetically allocated among various stakeholders).

\subsection{Evaluator Protocol} 
\textbf{Peer evaluators} (n=104) assessed only the artworks without knowledge of the creation category (AI-only vs. mixed media). Each evaluator rated a median of six images (M = 5.52, SD = 0.64) on perceived effort and artistic value (creativity, technical quality, and visual appeal, separated). Like creators, evaluators also made judgments about rights control and hypothetical value distribution(see evaluation interface in Figure~\ref{fig:interface}).

\noindent\textbf{Expert Evaluators} (n=5) Experts viewed all 105 submissions, ranked their top six selections, and wrote statements describing their ranking criteria and reasoning.

\subsection{Measures}
\label{subsec:measures}
We collected both behavioral observations and self-reported perceptions, measuring effort, artistic value constructs (creativity, technical quality, visual appeal), hypothetical economic value, and rights allocations. Full survey instruments are provided in Appendix~\ref{app:survey}. This section overviews the key measures and their corresponding survey questions.

\emph{Control variables.} We measured several participant characteristics to control for potential confounds in our quantitative analysis (see Section\S\ref{sec:analysis_plan}: AI usage frequency (7-point scale from "Never" to "Daily"); for creators, display preferences (museums, commercial galleries, commercial reproduction, public spaces); for evaluators, museum comfort (7-point Likert scale for comfort with AI art in museums).

\subsubsection{Effort}

We measure effort through both behavioral observation and self-reports. For creators, the platform automatically logged three behavioral indicators: total prompt word count (sum of words across all prompts submitted), number of prompts (count of generation attempts), and session duration (total active minutes). These measures were highly intercorrelated (prompts $\times$ word count: $r = .63$; prompts $\times$ time: $r = .96$; word count $\times$ time: $r = .68$). Given this collinearity, we selected total prompt word count as our primary observed effort measure because it captures both iteration quantity and semantic elaboration. Responses were measured on a likert scale. All Likert-type responses were linearly rescaled to a common 5-point scale to enable consistent comparison and analysis across items.

\subsubsection{Artistic Value: Creativity}

We adopt a layperson's conception of creativity aligned with the ``Standard Definition''~\cite{Runco2012Standard}: the production of outputs that are both novel and appropriate. This differs from copyright law's lower ``originality'' threshold~\cite{feist1991} and patent law's higher ``novelty'' requirement \cite{usc35novelty}. We employed a single-item holistic measure rather than decomposing creativity into subdimensions. This approach is consistent with how creativity is typically assessed in authentic evaluative contexts: art juries, grant panels, and editorial decisions rely on integrated expert judgment rather than dimension-by-dimension scoring~\cite{kaufman2013furious}. Creators and evaluators rated the creativity of their artwork on a 1--5 scale.

\subsubsection{Artistic Value: Technical Quality}

Technical quality reflects execution fidelity, craftsmanship, and the effectiveness of combining algorithmic outputs with human curation~\cite{Kozbelt2004, rapp_how_2025}. Creators and evaluators rated the technical quality of their artwork on a 1--5 scale.

\subsubsection{Artistic Value: Visual Appeal}

Visual appeal captures aesthetic properties and perceived beauty, which shape judgments of artistic value~\cite{Leder2004Model, Reber2004}. Only evaluators rated visual appeal on a scale of 1--7, which was rescaled to a 1--5 scale.

\subsubsection{Hypothetical Economic Value}

We measured hypothetical willingness-to-pay (evaluators) and willingness-to-accept (creators) using ordinal price bands, following standard practice in experimental aesthetics~\cite{Newman2012}, to capture perceived worth thresholds \cite{Carson2005}. Evaluators responded to questions on the amount they are willing to pay, and Creators responded to questions on the amount they are willing to sell for on cost scales of \$0 (would not purchase/sell), less than \$50, \$51--\$100, \$500, \$1000, and \$3000 or above. Although the economic valuations collected by our study remained hypothetical, our setting allowed us to observe how judgments shift when stakeholders engage with AI art in authentic curatorial contexts~\cite{Carbon2020_EcologicalArt, Darda2025_MuseumVsDigital}; we discuss the relationship between stated and revealed preferences further in \S\ref{sec:limitations}.
 
\subsubsection{Hypothetical Value and Rights Allocation}

Participants hypothetically allocated monetary rewards and exclusive rights among five stakeholder categories, following prior work on attribution in human-AI co-creation~\cite{Epstein2020_WhoGetsCredit, lima_public_2025}: (i) the user/creator who prompted the AI, (ii) training data artists whose works were used to train the model, (iii) technical entities (the AI model, developers, or company), (iv) traditional toolmakers (e.g., Photoshop, brushes), and (v) v no one. Each allocation was binary (allocated = 1, not allocated = 0), and participants could select multiple recipients. Questions asked who \textit{should} receive rewards and rights, eliciting normative judgments about fair distribution rather than beliefs about current legal entitlements.

\subsection{Recruitment}
\label{sec:participants}
We recruited 253 university students as creators and peer evaluators via several university channels (university lists, university courses, campus advertisements, and maker communities). Participants signed up via a survey that collected their informed consent, demographic information, field of study, experience with AI models, and computer science, art, and law training and experience. Participants could choose to participate either as creators or peer evaluators; all were initially paid \$15 to keep monetary incentives constant. Failure to follow guidelines or incomplete responses excluded 44 participants, leaving 209 final participants: 105 creators and 104 evaluators. 

The sample comprised 70.8\% undergraduates, 12.4\% master's students, 3.3\% PhD students, and 13.4\% part-time or non-students. Demographics included 41.6\% Asian, 6.7\% Black or African American, and 38.3\% White participants. 60.3\% of participants were women, 37.3\% were men, and 2.4\% were non-binary or did not disclose their gender.

\paragraph{Disciplinary training.} We classified participants by self-reported training:

\begin{enumerate}
    \item Artistic training (26.2\%): Formal coursework in visual arts, arts-related training programs, or current/past art-related occupation. This broad definition captures varying expertise levels, from students with introductory coursework to practicing artists.
    
    \item Computer science (CS) training (23.6\%): CS coursework, degree programs, or experience building AI/ML models.    
  
    \item Legal training (19\%): Law students or legal professionals. We note this category does not specifically indicate intellectual property expertise; IP law is a specialized subfield, and general legal training may not confer detailed copyright knowledge.
    \end{enumerate}

Expert evaluators ($n$ = 5) were recruited through the authors' professional networks. They had an average of 20 years of experience in fields greatly affected by generative AI, including art curation, exhibition design, film/animation production, and creative journalism. They also had different expertise: two were practicing artists with international exhibition experience, one a film director, one an art faculty member, and one an arts journalist. We collected qualitative data on their perceptions of the artworks and rationale on each artwork.

\subsection{Analysis Plan}
\label{sec:analysis_plan}

We employed a mixed-methods approach combining quantitative analysis of survey responses and behavioral logs, with qualitative analysis of open-text process descriptions and expert juror evaluation statements. Our analysis is primarily exploratory, examining how participants perceive and allocate value, authorship, and rights. We analyze participant groups (creators, peer evaluators) separately rather than making direct statistical comparisons given the different question phrasing between groups detailed in $\S$\ref{subsec:measures}.

\subsubsection{Quantitative Analysis}
We used regression modeling to examine relationships between value perceptions, rights allocation, and participants' disciplinary training, in line with our research question. We report coefficients ($B$) for linear and odds ratios (O.R.) for logistic models, 95\% confidence intervals for those coefficients, and $p$ values ($\alpha$<0.05) for results discussed in the text. We include full regression tables in the Appendix.

\begin{itemize}
    \item \textbf{RQ1:} To investigate artistic value, we examine the relationship between one sub-component of artistic value (i.e., creativity) and the other sub-components (i.e. technical quality and visual appeal). This initial analysis reveals their intercorrelations or baseline relationships that guide the interpretation of subsequent models, where each sub-component serves as a dependent variable. We then examine the relationship between these sub-components and economic value. We controlled for potential confounds in all regressions: AI usage frequency and acceptable display contexts (for creators, where they would prefer to display their art; for evaluators, acceptability of displaying AI art in museums).
    
    \item \textbf{RQ2:} To examine the distribution of rewards between stakeholders, for each participant group, we constructed three logistic regression models with the dependent variable being whether participants would allocate monetary rewards or exclusive rights to (i) the user/creator, (ii) training data artists, or (iii) technical entities. Independent variables included creativity, effort, technical quality, visual appeal (for evaluators only), AI usage frequency, display preferences, and training. We do not analyze allocations to traditional toolmakers or ``no one'' due to low base rates (toolmakers: 2.67\% creators, 5.34\% evaluators; no one: 3.23\% creators, 1.02\% evaluators). The regressions were designed to precisely test how the perception of artistic values influences reward distribution.
    
    \item \textbf{RQ3:} Disciplinary training (art, CS, law, as defined in Section~\ref{sec:participants} was included as an independent variable in all models to examine how disciplinary training shapes value perceptions and attribution patterns.
\end{itemize}

\emph{Model Specifications.} For evaluator data, we used mixed-effects models with random intercepts for evaluators to account for multiple ratings. For creator data (one observation per participant), we used standard linear or logistic regression. Following established practice for multi-category ordinal scales, we treated Likert responses as continuous in linear models~\cite{Norman2010, Sullivan2013}~\footnote{We also conducted robustness checks. Specifically, we reran the analysis treating the Likert scales as strictly ordinal data using ordinal logistic regression. This complementary analysis yielded identical patterns of significance and directionality}. 

\emph{Variables and Controls.} All continuous variables were rescaled for interpretability. Variables originally on scales other than 1--5 were rescaled to 1--5 (1 = None/Not at all to 5 = A lot/Extremely).

\subsubsection{Qualitative Analysis}
We conducted thematic analysis~\cite{BraunClarke2006} of creator open-text responses (process descriptions, inspiration sources, challenges, skills utilized) and expert juror written rationales. We use qualitative findings to interpret quantitative effects by grounding statistical patterns in participants' described experiences.

\paragraph{Coding Procedure.} First-cycle open coding applied descriptive labels (e.g., ``iteration/refinement,'' ``prompt engineering''). We ran a second-cycle coding to group related codes into higher-order themes, see codebook in Appendix~\ref{app:codebook}. Two coders completed 90-minute training, pilot-coded a stratified 10\% sample, and locked the codebook before main coding~\cite{MacQueen1998}. Double-coding of a random 20\% sample achieved acceptable inter-rater reliability (Krippendorff's $\alpha \geq 0.67$ per theme)~\cite{Krippendorff2011Alpha}.

\paragraph{Expert Juror Analysis.} We applied the same codebook to analyze expert jurors' written rationales for their selections. We use qualitative findings from expert jurors to contrast expert criteria with peer evaluator judgments, highlighting convergences (e.g., emphasis on creativity) and divergences (e.g., how technical quality relates to rights attribution).

\paragraph{Reflexivity.} Both coders had trainings in arts and computer science. We discussed how this might shape our interpretation of themes like ``prompt engineering as skill'' and actively sought counterexamples that pushed back on our assumptions~\cite{mcdonald2026reflexivity}.

\subsection{Exhibition}
\label{subsec:exhibition}
The expert and peer jurors co-curated the exhibition: we aggregated juror picks (top 6) into a panel shortlist (ties resolved in a brief adjudication meeting) and combined this with the top 6 from the peer-evaluation shortlist, which was constructed by ranking summed ratings of creativity, visual appeal, and technical quality to obtain the top 12 works for public display. 150 people attended the exhibition opening. 

\subsection{Research Justice \& Reproducibility}  

We recognize that experiential work on AI art intersects with sensitive issues of appropriation and creative rights. As steps toward research justice, we donated matching funds equivalent to the AI costs of this research to the Concept Art Association, an organization that advocates for artists and creators. In addition, we purchased carbon credits to offset the estimated energy use of our AI experiments.

\paragraph{Research Positionality.} Our research team includes members with education and professional experience in arts, law, and computer science. This interdisciplinary composition shapes our framing of AI art as simultaneously a creative, legal, and technical phenomenon. We acknowledge that our disciplinary training may influence how we interpret creator-evaluator results: legal training attunes us to rights and ownership frameworks, while computer science training emphasizes model development and user experience, and arts training sensitizes us to questions of authorship, creative labor, and aesthetic judgment. We attempted to mitigate disciplinary bias through collaborative interpretation and by grounding claims in participant language.

\paragraph{Ethical Considerations.} Beyond IRB approval, we addressed several ethical concerns specific to AI art research. Creators provided consent for (1) research participation and (2) public exhibition of their work. Creators retained the right to withdraw their work from the exhibition at any time. We discussed with creators how their artistic statements would be edited for wall labels and obtained approval before display. We acknowledge the need for research justice to account for the impacts of using generative AI tools on artists and on the environment. Matching funds equivalent to the AI-related exhibition research costs was donated to organizations advocating for artists’ and creators’ rights and carbon offset credits equivalent to the environmental impact of our research was purchased.

\subsection{Limitations}
\label{sec:limitations}

Our study has limitations that should guide the interpretation of findings. Participants were drawn from a single university community, which is comprised predominantly of undergraduates (70.8\%), limiting generalizability to professional artists, established legal practitioners, or general populations. While we classified participants by disciplinary training, these categories reflect training rather than expertise. 

Our study was also observational rather than experimental: we did not manipulate variables and thus cannot make causal claims. Unlike prior work that directly compared human-made versus AI-generated artworks~\cite{chamberlain2018putting, van_hees_human_2025, gangadharbatla2022ai}, we focused exclusively on AI-involved art, precluding direct comparison of how participants would value or allocate rights for purely human-made art. Additionally, unlike prior studies that focused on process manipulation in both ideation and implementation stages~\cite{messer_co-creating_2024, kirk2009modulation}, no process information was shown to evaluators.

Creator and evaluator questions differed in framing: creators rated skill/creativity \textit{use} (process-focused) while evaluators rated perceived \textit{quality} (product-focused). While theoretically motivated, this precludes direct comparisons between groups. We thus analyze groups separately. Similarly, our primary behavioral effort measure (prompt word count) may index verbosity rather than skilled creative labor. While the exhibition context was authentic, pricing measures were hypothetical stated preferences rather than revealed preferences~\cite{miller2011hypothetical}. We describe findings as ``hypothetical economic preferences'' and interpret patterns as indicative rather than predictive.

Questions about allocation asked who should receive rewards, which participants may interpret normatively (as ethical questions about deservingness) or descriptively (as predictions about current market/legal outcomes). Our measures allow us to examine correlates that may be associated with these different interpretations, though we do not claim that it directly identifies participant's underlying orientation. For example, we added prior AI experience, usage frequency, and training may make participants more attentive to existing production practices and market conventions, whereas perceived creativity and effort may be more closely associated with perceptions of who deserves the rewards. By including both sets of predictors in the regression, we examine whether reward allocation is more strongly associated with participant's AI usage frequency or with their evaluations of the artistic contribution itself.

We address this dual interpretation by using our independent variables to isolate these mindsets. Participants acting descriptively should be heavily influenced by their AI usage frequency and prior training, as these factors reflect real-world market knowledge. Conversely, participants acting normatively should place higher weight on intrinsic artistic value like creativity and effort. By including both sets of variables in our regression, our model controls for these competing mindsets and allows us to see which orientation exerts a stronger influence on reward allocation.

\begin{figure}
    \centering
    \vspace{-10pt}
    \includegraphics[width=0.95\linewidth]{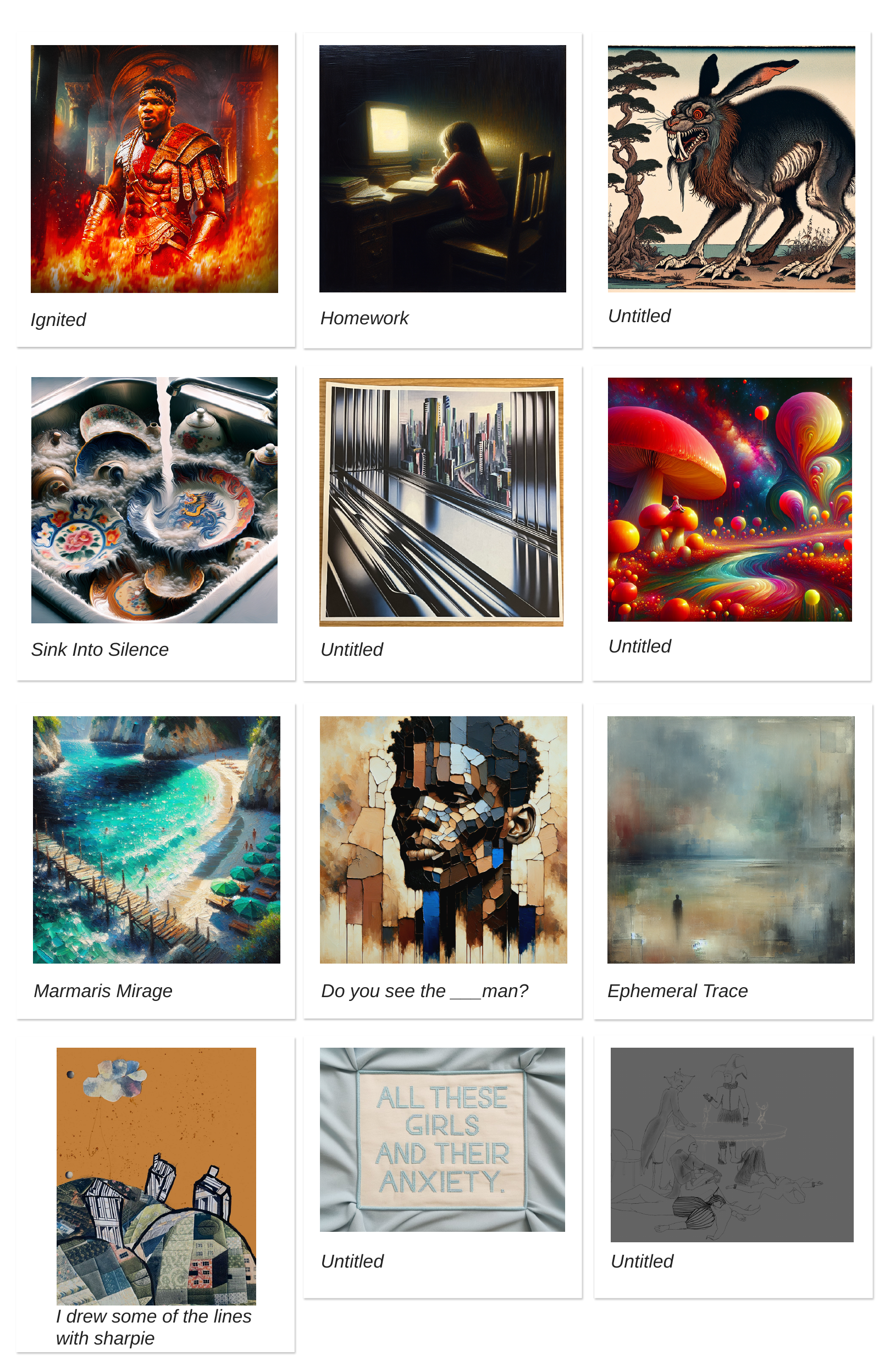}
    \caption{The twelve highest-rated artworks selected for the in-person public exhibition. Titles were chosen by the creators.}
    \label{fig:top-images}
    \Description{top rate artworks}
\end{figure}

\section{Results}
Here, we first contextualize our results ($\S$\ref{sec:creative_process}) by offering background on the process through which the artwork evaluated in this study was created (see sample artworks generated in Figure~\ref{fig:top-images}). Next, we examine how participants valued artworks artistically and economically (RQ1, $\S$\ref{sec:rq1}) and how they attributed rewards and rights among stakeholders (RQ2, $\S$\ref{sec:rq2}). Throughout, we demonstrate how disciplinary training in art, computer science, and law shaped these judgments (RQ3).

\subsection{Creative Process}
\label{sec:creative_process}
Creators used a median of 12 prompts before submitting their final work, in addition to any use of traditional or digital art methods among those who entered into the mixed-media category (mean = 20.65, standard deviation \textit{SD} = 28.92). Creators averaged 880.5 total words across all prompts before making their submission (\textit{SD} = 2422.44 words). Participants spent an average of 40.88 minutes (\textit{SD} = 53.03) creating. Measures of observed effort were highly co-correlated: number of prompts and total prompt word count ($r$ = .63, $p < .001$), number of prompts and active time creating ($r$ = .96, $p < .001$), and prompt word count and active time ($r$ = .68, $p < .001$). Given these correlations, we selected prompt word count as our observed effort measure because it captures both the number of prompts (e.g., design iterations) and the semantic richness of creative instructions.

Creators consistently described their process as one of iterative refinement. Some creators generated their artwork \textit{``by persistently refining my direction, adjusting prompts carefully, and focusing on what AI could do best while ensuring that my artistic vision remained in control''} (P4). Others worked more collaboratively with the system, even requesting advice from the system on how to best leverage it in creating their artwork. For example, P13 describes prompting the system to generate guidance on the creative process: \textit{``I first submitted the following prompt [to the system]... `Acting as an artist who is looking for creative ways to utilize AI image generators in order to submit a piece for an AI Art Competition, propose five ideas for prompts which can be given to the DALL-E AI Image Generator...' ''} Others first tested the system's functionality before coming up with their own ideas for what to create: \textit{``I began by writing a few random prompts to see how images were generated. Then, I started brainstorming various ideas for the artwork''} (P5).

\subsection{How Do People Value AI-Involved Artistic Works Artistically and Economically(RQ1/RQ3)}
\label{sec:rq1}
We found a significant correlation between perceptions of the artistic value (creativity, visual appeal, technical quality) of GenAI art and the perceived or observed effort of generating the piece; this is consistent with prior work on evaluations of visual art made with traditional techniques~\cite{Kozbelt2004, Hagtvedt2008Art}. Consistent with prior work in an online setting from which we adapt the method~\cite{lima_public_2025}, creators rate their works as more creative and requiring more effort than evaluators (Figure~\ref{fig:artistic_dimensions}, while evaluators perceive the works as having more technical quality (Figure~\ref{fig:artistic_dimensions}). These perceptual differences translate into differential pricing and reward distribution. Evaluators give price premiums for works they perceive as more visually appealing, creative, and effortful, while creators' asking prices are associated with their own evaluations of effort. Disciplinary training relates to these value decisions: those with legal training support are more likely to compensate training artists, while those with training in computer science are less likely to support creator compensation.

\subsubsection{Artistic Value}

\begin{figure}[t]
    \centering
    \includegraphics[width=\linewidth]{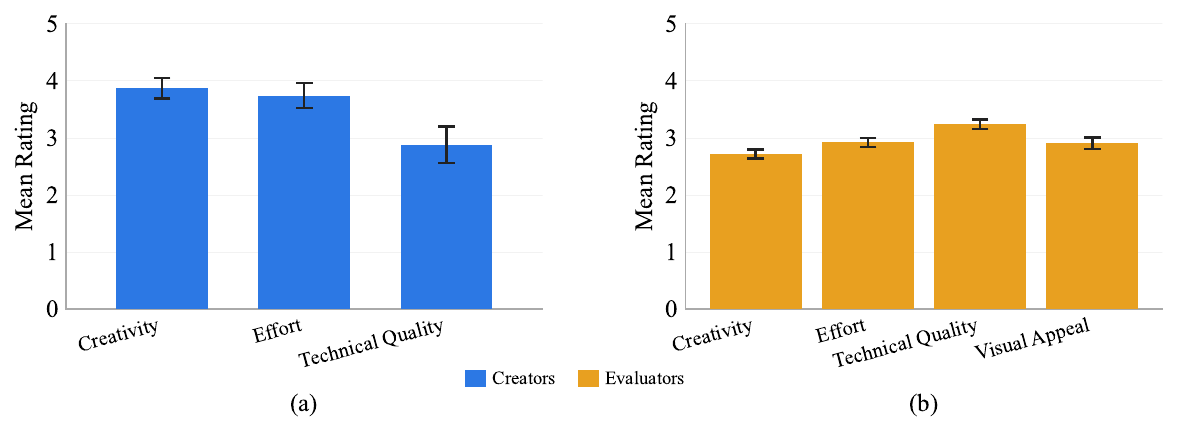}
    \caption{Mean ratings of perceived effort and artistic value dimensions by creators and evaluators. (a) Evaluators rated creativity (mean = 3.59), technical quality (mean = 4.34), visual appeal (mean = 3.86), and effort (mean = 3.88). (b) Creators rated their own work on creativity (mean = 3.86), technical quality (mean = 2.90), and effort (mean = 3.74). All ratings are on a 1---5 scale. Error bars represent 95\% confidence intervals}
\label{fig:artistic_dimensions}
\Description{Mean ratings of effort and artistic value}
\end{figure}

Among the three artistic values, evaluators rated technical quality most favorably (Figure~\ref{fig:artistic_dimensions}), with a median rating of ``moderate'' (mean = 4.37, \textit{SD} = 1.5). Their median perceptions of the works were that they were ``somewhat appealing'' (visual appeal: mean = 3.86, \textit{SD} = 1.8) and ``somewhat creative'' (creativity: mean = 3.58, \textit{SD} = 1.63). Consistent with prior findings that creativity, technical quality, and visual appeal jointly drive judgments of traditional art~\cite{Kozbelt2004, Hagtvedt2008Art}, we observed that these factors were highly correlated. Specifically, evaluators’ assessments of creativity were positively associated with their perceptions of visual appeal ($B = 0.25$ [0.17, 0.33], $p < .001$; see Appendix~\ref{app:creativity_perceptions}) and technical quality ($B = 0.17$ [0.07, 0.26], $p < .001$). 

Evaluators' median perception was that ``a little effort'' had gone into creating the works they evaluated (mean = 3.88, \textit{SD} = 1.46). Their perceptions of creativity were strongly correlated to their perception of effort ($B = 0.43$ [0.34, 0.52], $p < .001$). We found no significant relationship between disciplinary training and evaluators' perceptions of artistic value.

Creators on the other hand felt they put ``substantial'' creativity (mean = 3.87, SD = 1.23
) and effort (mean = 3.76, SD = 1.19
) into their submission. They perceived the technical quality of their artwork as ``moderate'' (mean = 2.89, $s = 1.18$, 
near the scale midpoint. Creators' perceptions of the creativity of their work were correlated with their perception of its technical quality ($B = 0.25$ [0.06, 0.43], $p < .011$), and also with observed effort as measured by total prompt length across all iterations ($B = 0.42$ [0.05, 0.79], $p < .026$). Training in computer science marginally increased perceived creativity ($B = 0.39$ [-0.06, 0.84], $p < .090$).

In describing how they created their artwork, participants frequently emphasized the effort and skill involved in the process. Participants themselves linked effort to the impression of creativity. For example, P08 noted that meaningful art requires \textit{``effort, attention to detail, skill, and sheer talent,''} while P09 explained that AI art has value \textit{``because it requires some level of human artistic skill and a greater degree of effort to create.''} Such reasoning echoes the historical ``sweat of the brow'' doctrine, which linked labor to creative rights but was later rejected in U.S. copyright law~\cite{feist1991}.

The preceding analyses reveal that creators who invested more effort perceived their work as more creative, while evaluators' creativity judgments were associated with their \textit{perceptions} of effort. But do evaluators accurately perceive the effort creators actually invested? To test this, we examined whether creators' observed behavioral effort (total prompt word count) correlated with evaluators' perceptions of the same artworks. Creator effort did not significantly correlate with evaluator-perceived effort ($B = 0.01$ [$-0.16, 0.18$], $p = .899$), perceived creativity ($B = -0.03$ [$-0.22, 0.15$], $p = .728$), visual appeal ($B = -0.05$ [$-0.26, 0.16$], $p = .651$), technical quality ($B = -0.03$ [$-0.19, 0.12$], $p = .681$), or willingness-to-pay ($B = -0.07$ [$-0.18, 0.04$], $p = .226$).

\begin{figure}[t]
  \centering
  \includegraphics[width=1\linewidth]{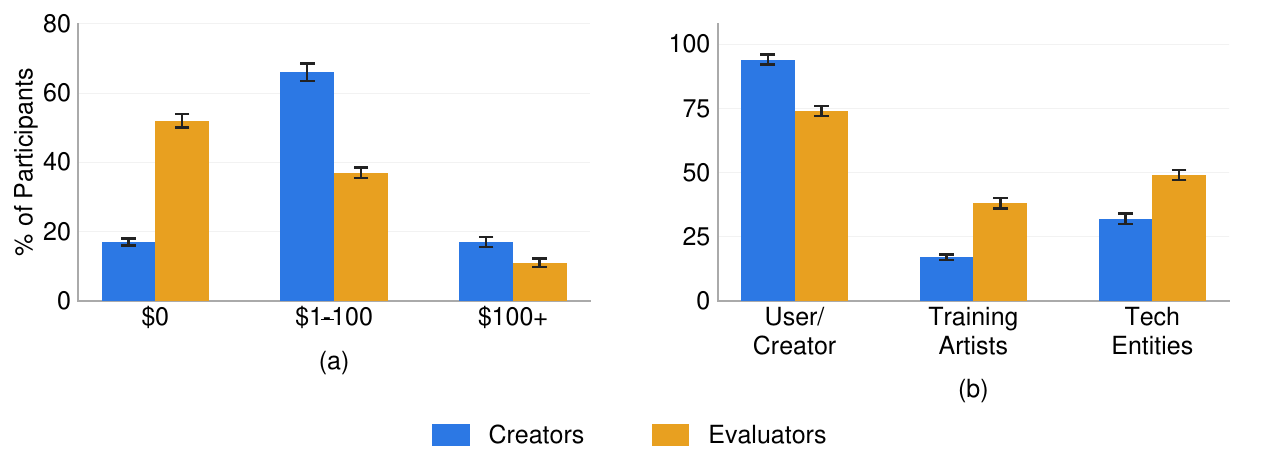}
  \caption{Valuation patterns and pricing preferences for AI-involved artwork by participant group. (a) Distribution of prices creators would sell for, and evaluators would pay, (b) Percentage of participants who would allocate monetary compensation to each stakeholder group. Error bars represent standard errors capped at 5\%.}
  \label{fig:monetary_prize}
  \Description{Valuation patterns and pricing preferences}
\end{figure}

\subsubsection{\textbf{Hypothetical Economic Value}} 

A little more than half of evaluators (54\%) stated they would not buy any of the art they evaluated, while 36\% indicated they would pay \$100 or less for at least one piece, and 10\% would pay above \$100 (Figure~\ref{fig:monetary_prize}). Among evaluators who reported being willing to buy, pricing was positively associated with perceived creativity (B = 0.14 [0.08, 0.20], $p < .001$, Appendix~\ref{app:price}), visual appeal (B = 0.19 [0.13, 0.25], $p < .001$), and effort (B = 0.13 [0.06, 0.20], $p < .001$). Evaluators with CS training reported less willingness to pay (B = -0.38 [-0.66, -0.10], $p < .008$).

More creators reported being willing to sell their works (71.2\%) than evaluators were willing to buy them (Figure~\ref{fig:monetary_prize}). 56.7\% of creators valued their work at \$100 or less, while 14.4\% would ask above \$100. Creators' asking prices were positively associated with observed effort (B = 0.64 [0.18, 1.11], p = .008). Those with training in creative art set lower asking prices (B = -0.58 [-1.13, -0.02], $p = .044$).

28.2\% of creators would give their work away for free. Participants' explanations for why also focus on effort, with some explaining that AI art lacks value because of its perceived accessibility---\textit{``it's too easy to create. I associate value with time, real artists dedicate lots of time and effort into their craft''} (P46)---and reproducibility---\textit{``anything that can be easily reproduced inherently has less value''} (P55). An additional 14.3\% of creators would not sell their work, either because they thought it had no value or because they personally valued it and did not want to give it up.

\subsection{How do people hypothetically distribute rewards and rights between stakeholders?(RQ2/RQ3)}
\label{sec:rq2}
We find that while both creators and evaluators support compensating creators, they differ in their overall attribution of rewards and rights. Creators focus primarily on self-compensation, while evaluators, especially those with legal training, adopt more distributed models, recognizing training data artists and technical entities, as well.

\begin{figure}[t]
  \centering
  \includegraphics[width=0.8\linewidth]{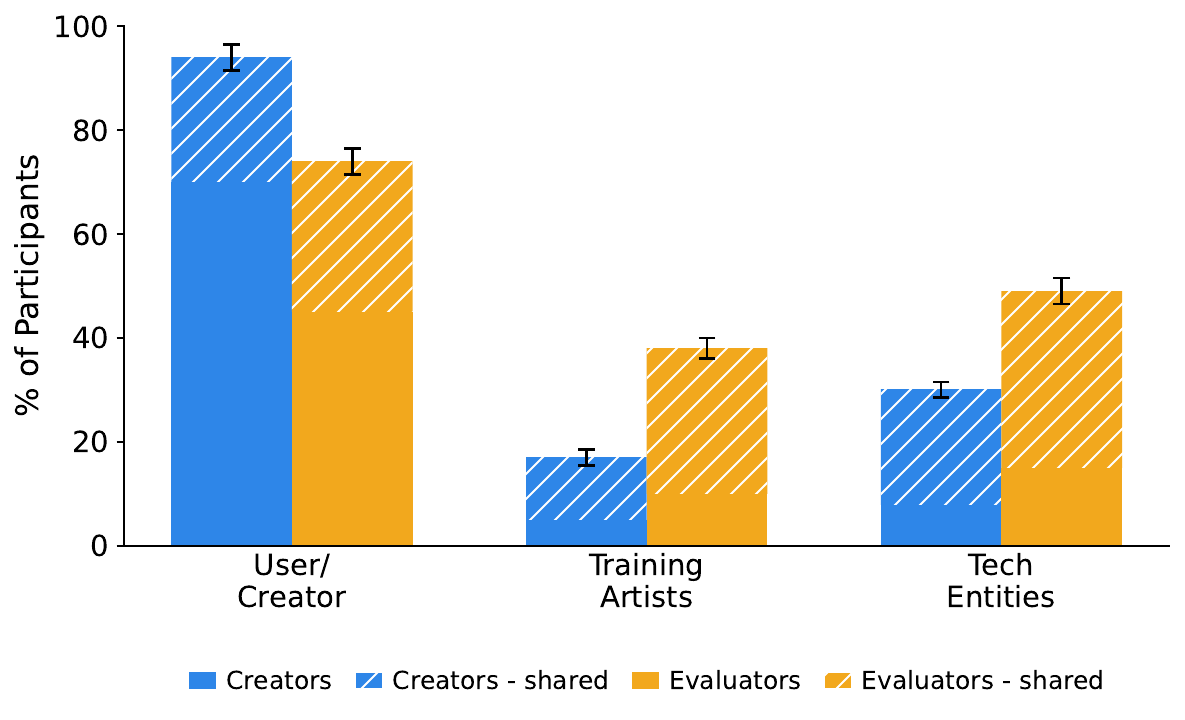}
  \caption{Monetary prize allocation preferences across stakeholder groups comparing creators and evaluators. Each pair of bars shows the percentage of creators (blue) and evaluators (orange) who selected each stakeholder group as a prize recipient. Solid colors indicate exclusive selection (only that recipient chosen), while striped patterns indicate shared selection (that recipient chosen along with others)}
  \label{fig:monetary_distribution}
  \Description{Monetary prize allocation}
\end{figure}

\begin{table*}[htbp]
\captionsetup{labelsep=colon}
\centering
\caption{Sample artworks with creator descriptions of their artistic process and inspirations. Creators drew on diverse sources, from childhood memories of museum artifacts (P95) to contemporary art references (P32), cross-cultural artistic fusion (P84), and personal cultural heritage (P61).}
\label{tab:sample_artworks}
\renewcommand{\arraystretch}{1.4}
\begin{tabular}{@{}>{\centering\arraybackslash}m{0.26\linewidth} >{\raggedright\arraybackslash}m{0.66\linewidth}@{}}
\toprule
Artwork & Artistic Statement \\
\midrule
\includegraphics[width=\linewidth]{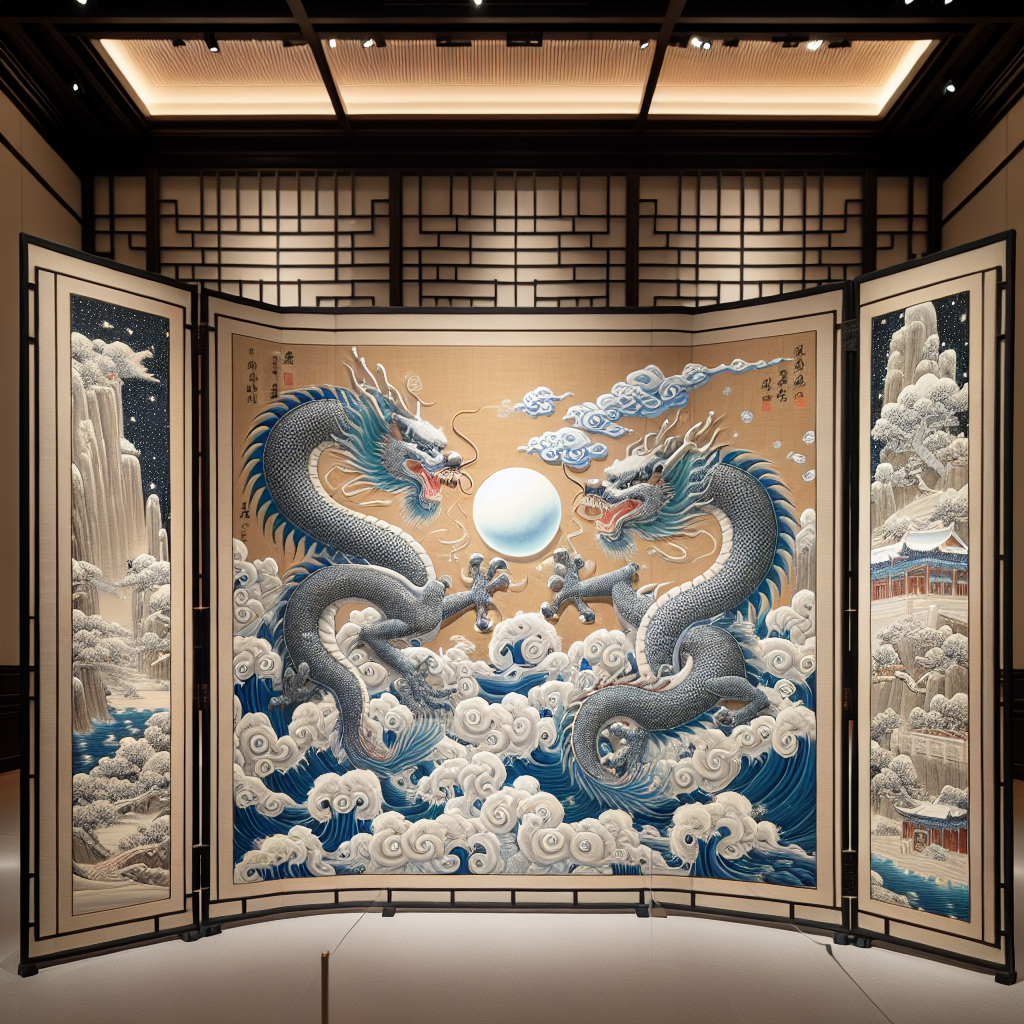} &
``It reminded me of a screen that used to be in the Smithsonian Museum of
Asian Art, which is no longer on display, and I cannot find records that it ever
existed. I realized with this tool I could recreate the screen to show other people
what it looked like and maybe help me find proof the real one existed. My
memory of the real screen itself is foggy, other than that it featured two dragons
playing with a ball in a snowy sky, and that it was Japanese, from sometime
between 1000--1500.''
\newline\mbox{}\hfill\textbf{--- P95} \\
\midrule
\includegraphics[width=\linewidth]{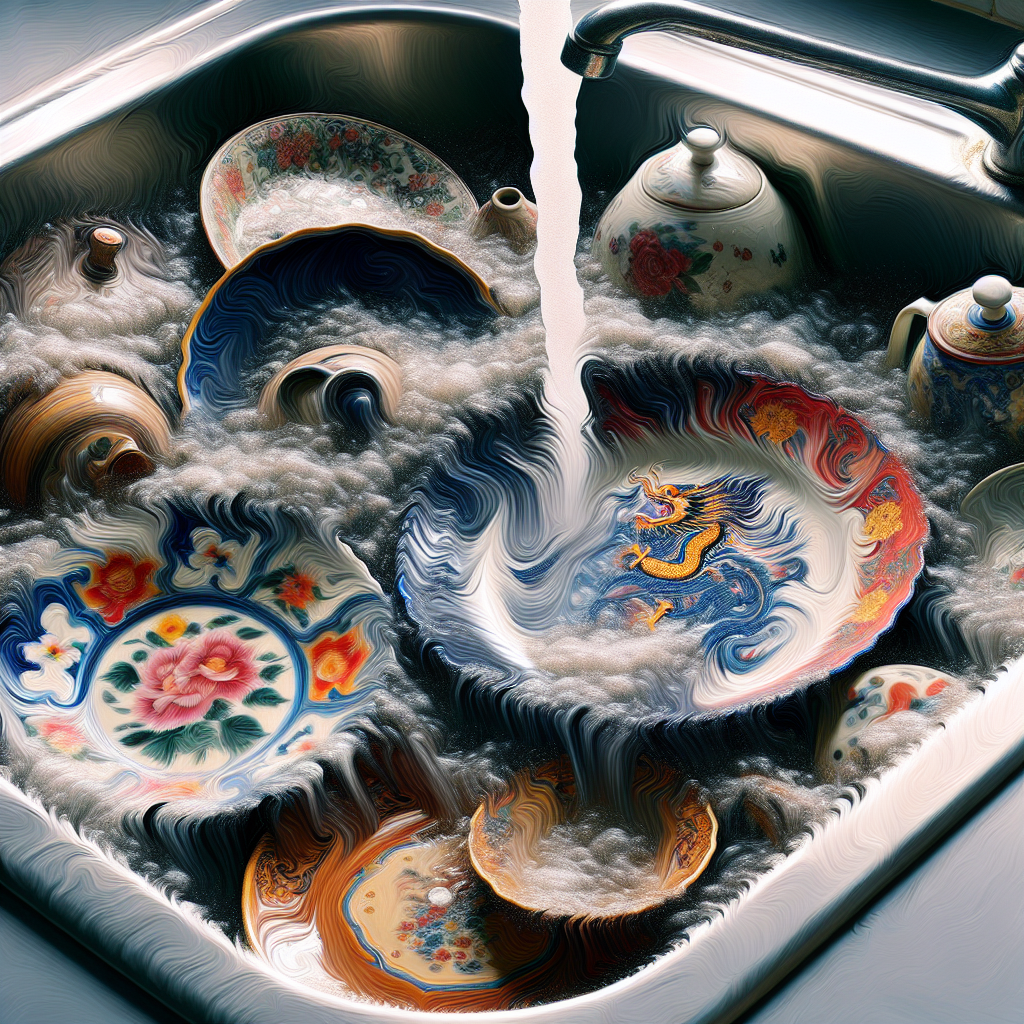} &
``I started with the idea of a dragon emerging from Peranakan-inspired plates
and then played with the idea of water hitting the plates, focusing on the flow of
the water symbolizing the `washing away' of culture and then settling on the
movement being a surrealist depiction of the plates themselves, withering,
washing or slipping away in a sink. I was inspired by a news article in a local
paper from home I saw recently about how younger generations tend to forget
or reject our culture.''
\newline\mbox{}\hfill\textbf{--- P61} \\
\midrule
\includegraphics[width=\linewidth]{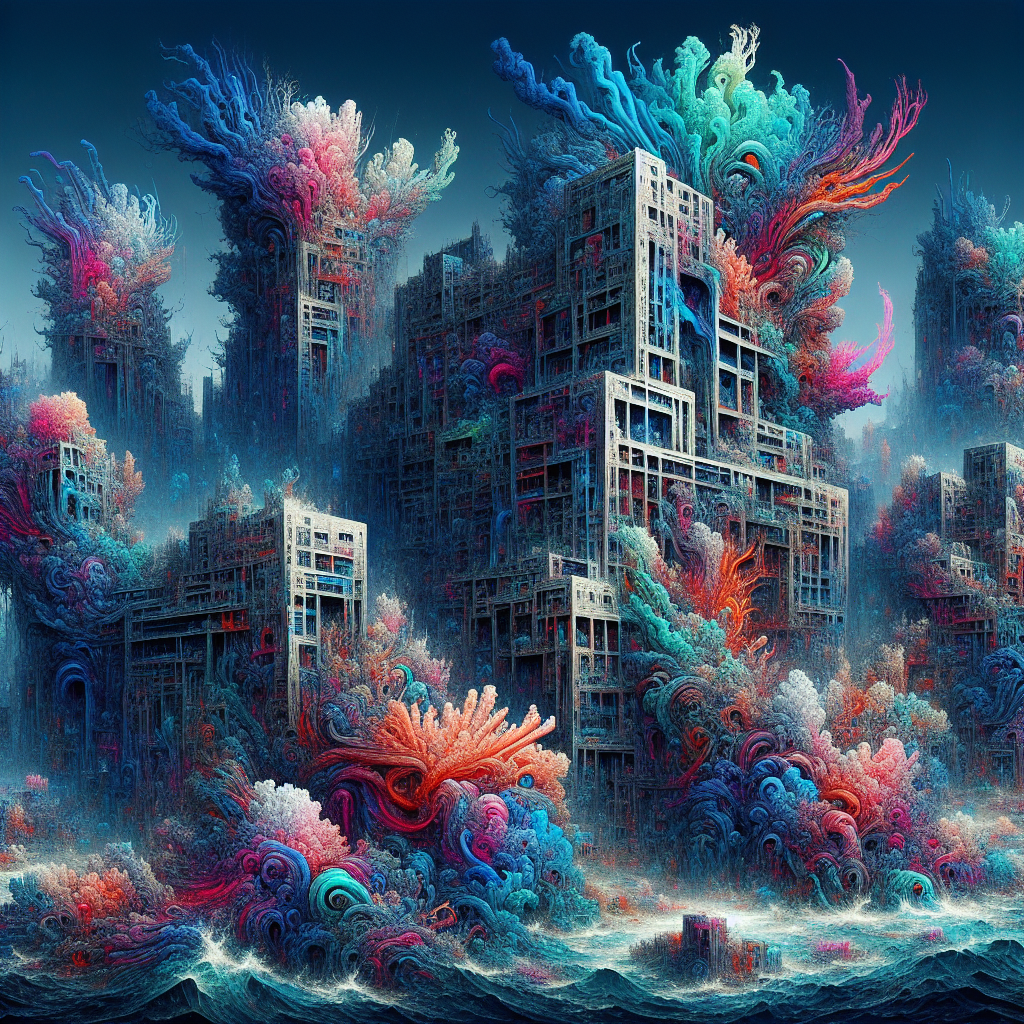} &
``I first started by combining two random interests of mine, coral reefs, and
brutalist architecture. Then I picked two of my favorite artists, Jean-Michel
Basquiat and Katsushika Hokusai, asking AI to weave in their artistic style to
the piece. The piece became more colorful, more interpretive, and the texture
slightly changed. The coral was overflowing in weird shapes and sizes just how
I wanted, and the style of Hokusai was present in the water around the piece.''
\newline\mbox{}\hfill\textbf{--- P84} \\
\midrule
\includegraphics[width=\linewidth]{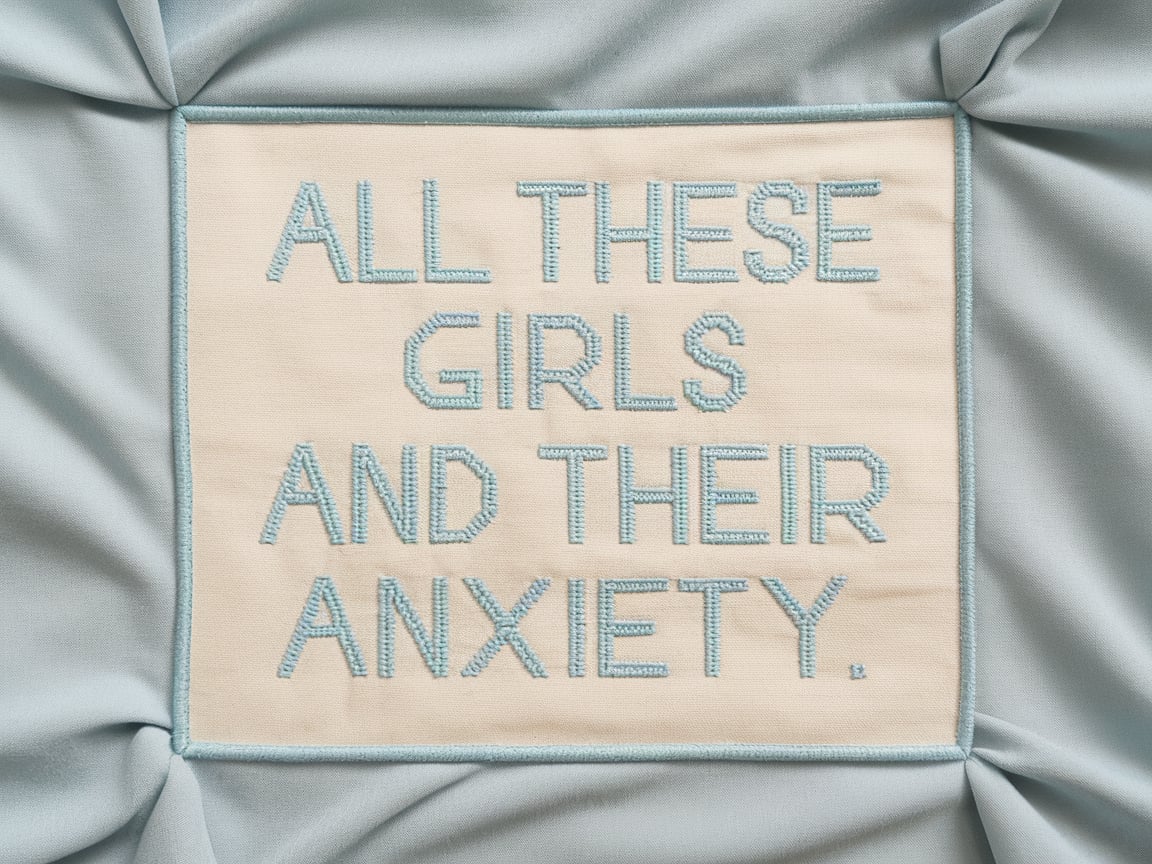} &
``I began by referencing the iconic Louise Bourgeois `I have been to hell and
back' poster. The most direct inspiration came from my friend's offhand
comment about `all these girls and their anxiety.' This phrase, delivered with
care and gentle humor, transformed my personal experience of anxiety into
something shared and less isolating. After generating the image, I had it printed
and framed, placing it in my bedroom where it would be the first thing I see
each morning---transforming my AI creation into a tangible daily affirmation.''
\newline\mbox{}\hfill\textbf{--- P32} \\
\bottomrule
\end{tabular}
\end{table*}
\subsubsection{Economic Compensation}

When considering who should receive some or all of the economic reward or monetary prize given by the exhibition, if awarded, the majority of creators (62.9\%) choose single beneficiaries, primarily themselves, while only 32.3\% chose to distribute between two or more groups, see Figure~\ref{fig:monetary_prize}.

All but 6 creators (94.3\%, Figure~\ref{fig:monetary_prize}) indicated that they would award themselves some or all of the prize. (62.9\% wanted all, 31.3\% wanted some).
Given the low variance in creator decisions, there are no significant covariates in the regression analysis of creators' self-compensation. Nearly three-quarters of evaluators would also award some or all of the prize to creators (39\% would award all, 35.5\% some). Evaluators who perceive a work as more visually appealing (OR = 1.34 [1.01, 1.77], $p < 0.040$) are more likely to allocate a monetary reward to its creator. However, those with training in computer science are less likely to do so than other evaluators (OR = 0.18 [0.05, 0.65], $p < 0.009$), perhaps reflecting a technical awareness of AI's contribution.

Creator and evaluator opinions diverged when we consider allocations to training data artists and technical entities. Few creators (18.1\%) would share rewards with training data artists. Among those who would share, one creator explained: \textit{``The credit and compensation for [the work's] development should partly go to the artists whose work the AI copied (likely without attribution)''} (P31). Some creators explicitly devalued their own work based on their awareness of the role of training data: \textit{``I do not think my AI art has financial value...given the original inspiration was drawn from major artists + copied''} (P12). Among the creators who would not sell their work, 31\% supported sharing rewards with training data artists, suggesting ethical concerns about derivative value. More evaluators would share rewards with training data artists (38\%), although still only a minority. Evaluators with legal expertise are significantly more likely to do so (OR = 9.18 [1.14, 73.57], $p<.037$), indicating a higher awareness of intellectual property issues and concerns related to derivative work. 

More creators (22.9\%) and evaluators (23.3\%) would share rewards with technical entities. Evaluators who perceive higher technical quality are significantly more likely to reward technology entities (OR = 1.63 [1.15, 2.32], $p < .006$), while perceiving greater human effort decreases these odds OR = 0.59 [0.41, 0.85], $p < .005$). This suggests evaluators may attribute technical quality to AI systems but effort-based value to human creators. Legal training also increased support for rewarding technical entities ($p < .042$), possibly reflecting understanding of innovative contributions.

\subsubsection{Rights Attribution}
Finally, we examined participants' views about who should hold legal rights and control over AI-involved artworks. This question extends beyond creative recognition to encompass legal ownership and commercial control.

We find that creators prefer to hold exclusive rights for their artwork (72.4\% vs. evaluators' 49.5\%). Creators with training in computer science are marginally more likely to want exclusive rights over their AI outputs (OR = 3.02 [0.89, 11.31], $p < .085$), while those who perceive their work as more creative were significantly more likely to support having exclusive rights (OR = 1.80 [1.08, 3.15], $p< .030$). Evaluators who perceive a work as requiring more effort to create (OR = 1.94 [1.38, 2.72], $p < .001$) or as more visually appealing (OR = 1.41 [1.05, 1.90], $p< .022$) are more likely to grant exclusive rights to the user. 

Relative to creators, evaluators are more likely to allocate exclusive rights to artists whose art was used to train the model, although only a minority of either group would do so (19.9\% vs 9.5\%). For creators, no variables significantly correlate with training artist rights attribution. For evaluators, a legal training significantly relates to this attribution (OR = 175.75 [2.69, 11479.59], $p<.015$), with perceived effort showing marginal influence (OR = 1.70 [0.97, 2.95], $p<.062$). More creators than evaluators would attribute exclusive rights to technical entities (23.8\% vs 10.8\%), although again only a minority would do so. Among creators, no variables significantly correlate with tech entity rights attribution. For evaluators, consistent with correlations for allocating the monetary prize, higher perceived technical quality increases rights attribution to technical entities (OR = 1.82 [1.11, 2.99], $p< .018$), again suggesting evaluators associate technical quality with the contributions of the AI model, developers, and company.

\section{Discussion}
Here we discuss the implications of our findings and suggest future directions for research and implementation, organized by research question.

\subsection{Valuation of AI-Involved Art (RQ1)}

Creators and evaluators construct value differently. For creators, actual effort—measured by total prompt word count—was positively correlated with asking prices: those who invested more labor valued their work more highly. Evaluators' hypothetical purchase prices were instead correlated with their perception of the effort invested in the work, as well as the artistic value they perceived the work to have. 

In traditional media, creative processes often leave material or informational traces that encode aspects of artistic labor. Paintings reveal brushstrokes that signal technique and stylistic decision-making~\cite{Wolfflin1950Principles, Gombrich1960ArtIllusion}; sculptural works bear tool marks indicative of physical effort; and software development preserves commit histories that document iterative labor~\cite{Kalliamvakou2014VersionControl}. Even digital photography retains EXIF metadata recording exposure and camera settings, providing residual evidence of production decisions~\cite{Ritchin2009AfterPhotography}. Such traces allow observers to infer creative investment even without direct access to the production process~\cite{Becker1982_ArtWorlds}.

AI-generated images, by contrast, appear instantaneous. The final artifact contains no embedded trace of the prompt refinement, selective regeneration, or iterative curation that shaped it~\cite{Oppenlaender2025}. As a result, the labor involved in AI-assisted creation becomes archaeologically unrecoverable from the output alone. When production processes leave no visible evidence, observers must rely on inference from surface features, which can systematically misrepresent underlying creative labor~\cite{Akerlof1970}.

Our findings mirror prior findings showing that creators’ valuations are shaped by their own creative investment, whereas external evaluators rely primarily on outcome-based judgments~\cite{BuccafuscoS2011CreativityEffect, kruger_effort_2004}. Evaluators in our study assessed only final artworks and were not shown process information. While this reflects typical viewing conditions, it prevents us from determining whether the observed mismatch is intrinsic to AI-generated outputs or instead a consequence of missing information. Prior work suggests that explicit disclosure of effort can increase perceived value~\cite{kruger_effort_2004}. However, prior work on AI-involved work finds that even when evaluators are provided with traces of the effort invested in an AI-involved work they perceive the effort invested as less than creators do~\cite{lima_public_2025}. We encourage future controlled experiments to distinguish between these possibilities. As part of such efforts, we encourage the development of new mechanisms that allow creators to document and communicate their creative process effectively.  For example, drawing on approaches such as model cards~\cite{Mitchell2019} and C2PA provenance standards~\cite{C2PAstandard}, AI-generation systems could generate structured records of iteration, curation decisions, and human guidance without exposing proprietary workflows.

Beyond market valuation, the invisibility of creative labor poses challenges for legal frameworks that require evidence of human creative contribution\cite{fritz_notion_2024, US_2025}. Future work to develop documentation standards that surface creative input could therefore support both fairer valuation practices and more robust legal recognition of human authorship.

\subsection{Distribution of Value  (RQ2)}

While creators and evaluators largely agreed that prompting users deserves monetary compensation, they diverged sharply on other stakeholders. Few creators were willing to share rewards with training data artists, while more evaluators were supportive—a pattern aligning with prior research suggesting evaluators are more willing to distribute credit across for AI-generated art~\cite{lima_public_2025}. Still, only a minority did so. Similarly, most creators strongly preferred exclusive rights for themselves, while more, but still a minority, of evaluators would distribute rights to training data artists. While evaluators are more distributive than creators, training data artists remain largely left behind in both groups' attribution frameworks.

A distributed view of creative contribution challenges conventional artistic notions of authorship, which have historically emphasized individual creative vision and direct execution as central to artistic identity~\cite{Becker1982_ArtWorlds}. While art has long recognized collaborative practices—from Renaissance workshops where master artists directed apprentice execution to contemporary artist collectives—these involve direct human co-creation with clearly delineated roles and contributions. AI collaboration introduces computational systems as intermediaries, trained on thousands of prior artworks without explicit consent or attribution~\cite{shi2023_understanding}, creating a fundamentally different form of distributed authorship that neither art traditions nor legal frameworks have fully addressed. While scholars continue to work on proposals for new authorship frameworks\cite{RamosZaga2025DistributedAuthorship, Goodfellow2024DistributedArtAuthorship, Fritz2025AuthorshipAI}, the distributed attribution preferences we observe can support these frameworks. Particularly evaluators' greater willingness to recognize training data artists, suggest that platforms should take a first step toward distributive norms by supporting flexible attribution models capable of acknowledging multiple contributors.

\subsection{Disciplinary training Shapes Attribution (RQ3)}

Disciplinary training systematically influences valuation and attribution. Evaluators with legal training were significantly more likely to reward and give control to training data artists. This pattern suggests that formal legal education sensitizes individuals to derivative-work concerns and intellectual property lineage~\cite{grimmelmann_copyright_2017}. 

By contrast, evaluators with computer science training were less likely to report wanting to compensate creators 
Suggesting that technical training, which may include information on how AI generation works technically, may lead to attributing greater creative agency to the model.
Finally, training in art was associated with lower asking prices among creators. Creators with formal art training set lower asking prices, perhaps reflecting greater familiarity with market norms or skepticism about the commercial value of AI-involved art. 

These systematic differences suggest that future professionals may bring divergent intuitions to policy debates. Legal professionals may advocate for distributed attribution models recognizing training data artists. CS professionals, likely understanding the system's technical workings, may reduce support for prompting users as primary contributors, possibly recognizing that the model and training data play substantial generative roles. Understanding these disciplinary patterns is critical for anticipating how stakeholder perspectives may be weighted in policy design.
We encourage future work exploring more specifically and experimentally what disciplinary content shapes beliefs about the role of different stakeholders in human-AI collaboration and how.

\subsection{Design Implications}

\label{app:design-implications}

Current image generation systems rarely produce durable records of the iterative process that shaped an output, including the prompts submitted, the generations rejected, and the pivots and refinements that our data show are central to creators' own sense of value. One possible approach could be provenance standards such as C2PA and structured process documentation modeled on model cards~\cite{C2PAstandard}. Model cards already provide technical infrastructure for surfacing prompt iteration depth, curation decisions, and evidence of human creative input without exposing proprietary workflows or creative strategies~\cite{Mitchell2019}. Platforms that implement these standards at the point of generation, rather than leaving documentation to creators after the fact, could enable fairer market valuations.

Existing provenance standards could also perhaps provide the kind of process evidence that legal frameworks increasingly require. If the labor that would justify authorship cannot be recovered from the output, existing legal doctrine has little foothold, regardless of how substantial the creative investment was. Documentation standards that surface process signals at scale, including prompt logs, session metadata, and generation histories, could provide the evidentiary basis that many AI-assisted works currently lack. Such standards may enable more consistent legal recognition without requiring courts to resolve deeper philosophical questions about machine creativity that remain unsettled across jurisdictions~\cite{Fritz2025AuthorshipAI, wipo2023_perception}. Our finding that legal training increases willingness to recognize training-data artists' contributions suggests that legal education may already sensitize practitioners to these concerns. Policymakers may therefore be more open to distributed attribution than current laws suggest.

More broadly, our results show that disciplinary background systematically shapes attribution judgments in ways that matter for system design. Evaluators with computer science training were significantly less likely to compensate the prompting user and more likely to attribute contribution to the model and its developers. This may reflect a technically coherent view of how much generative work diffusion models perform relative to the prompt specification. We do not claim that one group is right or wrong. Rather, we show that disciplinary homogeneity in system design can produce distributional consequences that should be made explicit~\cite{grimmelmann_copyright_2017}. The people building compensation mechanisms, attribution APIs, and governance frameworks are often CS-trained. As a result, the attribution intuitions most likely to be encoded in technical infrastructure may diverge from those of artists, legal experts, and the general public. Convening attribution and governance decisions across disciplinary lines, including artists and legal practitioners alongside engineers, is therefore a concrete step toward systems whose default behaviors reflect a broader range of stakeholder values.

\section{Conclusion}
We held an in-person, juried art exhibition to explore how creators and evaluators perceive value and rights in AI-assisted art. Legal and market frameworks often assume value and rights can be inferred from the work alone, but our data indicates that many creative actions important to stakeholder valuation and recognition are rendered invisible to external observers, particularly in GenAI art. This invisibility has consequences: stakeholders risk undervaluation, legal uncertainty, and misallocation of rights. Our findings suggest the need for systems and norms that help surface human creative input reliably through provenance tools\cite{moruzzi2025contentauthenticities}, standards for attribution, or documentation practices. Going forward, future work could test how interventions that make the creative process visible (e.g., disclosure of iteration history, prompt summaries) and cultural contexts outside the U.S. may impact perceptions, as well as explore how legal doctrine might adapt to distributed rights frameworks that recognize the creative labor of all who contribute to the creation of AI-involved art.


\bibliographystyle{ACM-Reference-Format}
\bibliography{ref}

\subsection{Additional overview of the System}
\label{app:system}
\textbf{Platform Architecture.} We developed a custom web-based AI art creation platform using React.js with Tailwind CSS for the frontend interface and Firebase services (Firestore database, Storage, and Authentication) for backend infrastructure. The system integrated directly with OpenAI's DALL-E 3 API for image generation at 1024$\times$1024 pixel resolution in PNG format with standard quality settings.
\\
\textbf{User Interface and Category-Specific Features.} Participants logged in using unique anonymized login IDs and could enter text prompts through the input interface. The platform supported two participant categories: (1) AI-Only category participants (loginId prefix: CAI25) could only use text prompts, with upload attempts triggering an informational alert, and (2) AI-Involved category participants (loginId prefix: HAI25) had additional capability to upload reference images up to 4MB in PNG or JPEG format. Generated images appeared in the main content area with all previous generations accessible through a sidebar history interface, and participants could download any generated image. An optional tutorial was available through the guidelines page, providing example prompts (``Generate an image of a cat in a comic-book style of art'' and ``Generate an image of a rabbit in an abstract style of art'') and explaining platform features, including the AI's smart memory capabilities that attempted to maintain consistency across generations within a session.
\\
\textbf{Data Collection and System Performance.} The system logged comprehensive data for research purposes including: (1) session information (login ID, start/end timestamps, duration, active time), (2) all prompts submitted with their timestamps and DALL-E 3's enhanced versions, (3) original uploaded images when applicable, (4) generated image URLs with associated prompts and timestamps, and (5) user interaction patterns such as generation history access and download actions. All data transmission used HTTPS/TLS encryption, participant identifiers were anonymized before analysis, and data handling complied with IRB protocols, with access restricted to research team members. The platform achieved over 99.5\% uptime with average generation times of 15--30 seconds, page load times under 2 seconds, and API success rates exceeding 98\%. Quality control measures included identical DALL-E 3 parameters for all participants, consistent API versioning, automated data validation, pilot testing, and continuous error monitoring.

\subsection{\textbf{Ethics and Limitations.}} Participants were informed about the use of DALL-E 3, data collection procedures, and session tracking for research purposes, and they retained rights to their final artworks per IRB-approved study protocol. The platform experienced occasional challenges, including DALL-E 3 content policy rejections for certain prompts, Firebase connection errors requiring retry attempts, image upload restrictions, and session timeouts after extended inactivity, with error handling implemented to log failures while providing user-friendly recovery feedback.

\subsection{Survey Questions}
\label{app:survey}

\subsubsection{Creator Submission Form}

\paragraph{Part A: Creation Experience}
\begin{enumerate}
  \item Please provide a detailed description of how you created your artwork. [Open text]
  \item Please describe how you used AI in your creation process. [Open text]
  \item Please describe how you (and/or other people) were involved in your creative process. [Open text]
  \item Please list the decisions you made as part of your creative process. [Open text]
  \item What influenced your final art piece (e.g., other art/artists, personal experiences)? [Open text]
  \item What difficulties did you encounter and how did you overcome them? [Open text]
  \item Considering your creative process, would you say your final artwork was mostly created by:
    \begin{enumerate}[label=(\alph*)]
      \item Completely human
      \item Mostly human
      \item Collaboration
      \item Mostly AI
      \item Completely AI
    \end{enumerate}
\end{enumerate}

\paragraph{Part B: Value and Skills Assessment}
\begin{enumerate}
  \item I used creativity to generate this artwork.
    \begin{enumerate}[label=(\alph*)]
      \item Scale: 1 (Strongly disagree) -- 7 (Strongly agree)
    \end{enumerate}
  \item I used technical skills (composition, colors, detail, etc.) to generate this artwork.
    \begin{enumerate}[label=(\alph*)]
      \item Scale: 1 (Strongly disagree) -- 7 (Strongly agree)
    \end{enumerate}
  \item I put in effort to generate this artwork.
    \begin{enumerate}[label=(\alph*)]
      \item Scale: 1 (Strongly disagree) -- 7 (Strongly agree)
    \end{enumerate}
  \item Do you believe your AI art has financial value? What factors influence this? [Open text]
  \item What skills did you use in this project? [Open text]
  \item How relevant was your educational background?
    \begin{enumerate}[label=(\alph*)]
      \item Not relevant
      \item Slightly relevant
      \item Moderately relevant
      \item Very relevant
      \item Extremely relevant
    \end{enumerate}
  \item Compared to other art projects, how much effort was required?
    \begin{enumerate}[label=(\alph*)]
      \item Much less effort
      \item Less effort
      \item Similar effort
      \item More effort
      \item Much more effort
    \end{enumerate}
  \item In which contexts would you want your artwork displayed or sold? (Select all)
    \begin{enumerate}[label=(\alph*)]
      \item Museums and traditional galleries
      \item Public spaces (e.g., subway stations, art fairs)
      \item Commercial reproduction (products, ads, digital platforms)
      \item Commercial art galleries
      \item Other [Open text]
    \end{enumerate}
  \item If you were to offer your piece for sale, at what price?
    \begin{enumerate}[label=(\alph*)]
      \item \$0
      \item Less than \$50
      \item \$51 to \$100
      \item \$500
      \item \$1000
      \item \$3000 and above
      \item I would not offer my piece for sale
    \end{enumerate}
\end{enumerate}

\paragraph{Part C: Authorship Questions}
\begin{enumerate}
  \item Who do you consider to be the author(s)? (Select all)
    \begin{enumerate}[label=(\alph*)]
      \item Me (the user of the AI model)
      \item The AI model itself (e.g., DALL-E)
      \item The company that developed the AI model
      \item The human developers of the AI model
      \item Makers of non-AI tools used (e.g., Photoshop, brushes)
      \item Artists whose works were used to train the AI model
      \item No one
    \end{enumerate}
  \item Why do you consider [selected entities] as authors? [Open text]
  \item Why don't you consider [non-selected entities] as authors? [Open text]
  \item If your art is a winning piece, who should receive the monetary prize? (Select all)
    \begin{enumerate}[label=(\alph*)]
      \item Me (the user)
      \item The AI model itself
      \item The company
      \item The developers
      \item Non-AI tool makers
      \item Training-data artists
      \item No one
    \end{enumerate}
    
  \item If multiple options selected: How should the prize be distributed? [Percentage allocation]
\end{enumerate}

\paragraph{Part D: Ownership Questions}
\begin{enumerate}
  \item Concerning rights to control this image, which applies?
    \begin{enumerate}[label=(\alph*)]
      \item Someone should have exclusive rights
      \item Anyone should be able to use it
      \item No one should control it (public domain)
    \end{enumerate}
  \item If exclusive rights selected: Who should hold them? (Select all)
    \begin{enumerate}[label=(\alph*)]
      \item Me (the user)
      \item The AI model itself
      \item The company
      \item The developers
      \item Non-AI tool makers
      \item Training-data artists
      \item No one
    \end{enumerate}
\end{enumerate}

\subsubsection{Evaluator Survey Questions}
\label{app:evaluator}

\paragraph{Part A: General Views on AI-Generated Art (Filled Once)}
\begin{enumerate}
  \item Do you feel the use of AI in creating artworks enhances or diminishes its value? 
  \begin{enumerate}[label=(\alph*)]
    \item Scale: 1 (Strongly diminishes) -- 7 (Greatly enhances)
    \item I don't know
  \end{enumerate}

  \item How comfortable are you with AI-generated art being exhibited in museums?
  \begin{enumerate}[label=(\alph*)]
    \item Scale: 1 (Very uncomfortable) -- 7 (Very comfortable)
    \item I don't know
  \end{enumerate}
\end{enumerate}

\paragraph{Part B: First Impression (Per Image)}
\begin{enumerate}
  \item To what extent do you think this artwork was created by humans or AI?
  \begin{enumerate}[label=(\alph*)]
    \item Completely human
    \item Mostly human
    \item Collaboration
    \item Mostly AI
    \item Completely AI
    \item I don't know
  \end{enumerate}
\end{enumerate}

\paragraph{Part C: Quality and Artistic Attributes (Per Image)}
\begin{enumerate}
  \item What is your impression of the creativity displayed in this artwork?
  \begin{enumerate}[label=(\alph*)]
    \item Scale: 1 (Not creative at all) -- 7 (Extremely creative)
    \item I don't know
  \end{enumerate}

  \item How would you rate the technical quality of the artwork (composition, colors, detail)?
  \begin{enumerate}[label=(\alph*)]
    \item Scale: 1 (Very low) -- 7 (Very high)
    \item I don't know
  \end{enumerate}

  \item How much effort do you think went into creating this artwork?
  \begin{enumerate}[label=(\alph*)]
    \item Scale: 1 (No effort at all) -- 7 (A great deal of effort)
    \item I don't know
  \end{enumerate}

  \item How would you rate the visual appeal of the artwork?
  \begin{enumerate}[label=(\alph*)]
    \item Scale: 1 (Not appealing at all) -- 7 (Very appealing)
    \item I don't know
  \end{enumerate}

  \item In which contexts should the artwork be displayed or sold? (Select all)
  \begin{enumerate}[label=(\alph*)]
    \item Museums and traditional galleries
    \item Public spaces (e.g., subway stations, art fairs)
    \item Commercial reproduction (products, ads, digital platforms)
    \item Commercial art galleries
    \item I do not think it should be displayed or sold
    \item Other [Open text]
    \item I don't know
  \end{enumerate}

  \item If this artwork were for sale, how much would you be willing to pay?
  \begin{enumerate}[label=(\alph*)]
    \item \$0 (I would not buy it)
    \item Less than \$50
    \item \$51 to \$100
    \item \$500
    \item \$1000
    \item \$3000 and above
    \item I don't know
  \end{enumerate}
\end{enumerate}

\paragraph{Part D: Authorship Questions (Per Image)}

\begin{enumerate}
  \item Who do you consider to be the author(s)? (Select all)
    \begin{enumerate}[label=(\alph*)]
      \item The person who used the AI model
      \item The AI model itself (e.g., DALL-E)
      \item The company that developed the model
      \item The human developers of the model
      \item Makers of non-AI tools used (e.g., Photoshop, brushes)
      \item Artists whose works were used to train the AI model
      \item No one
      \item I don't know
    \end{enumerate}

  \item Who should receive the monetary prize? (Select all)
    \begin{enumerate}[label=(\alph*)]
      \item The person who used the AI model
      \item The AI model itself
      \item The company
      \item The developers
      \item Non-AI tool makers
      \item Training-data artists
      \item No one
      \item I don't know
    \end{enumerate}

  \item If multiple selected: How should the prize be distributed? [Percentage allocation]
\end{enumerate}

\paragraph{Part E: Ownership Questions (Per Image)}

\begin{enumerate}
  \item Concerning rights to control this image, which applies?
    \begin{enumerate}[label=(\alph*)]
      \item Someone should have exclusive rights
      \item Anyone should be able to use it
      \item No one should control it (public domain)
      \item I don't know
    \end{enumerate}

  \item If exclusive rights were selected: Who should hold them? (Select all)
    \begin{enumerate}[label=(\alph*)]
      \item The person who used the AI model
      \item The AI model itself
      \item The company
      \item The developers
      \item Non-AI tool makers
      \item Training-data artists
      \item No one
      \item I don't know
    \end{enumerate}
\end{enumerate}

\begin{table*}[h]
\captionsetup{labelsep=colon}
\centering
\scriptsize
\renewcommand{\arraystretch}{0.9}
\setlength{\tabcolsep}{3pt}
\setlength{\heavyrulewidth}{1.5pt}
\setlength{\lightrulewidth}{1.0pt}
\caption{Literature Comparison. We compiled the most related works, including our prior work, to highlight the unique contributions of this study.}
\label{tab:literature}
\begin{tabular*}{\textwidth}{@{\extracolsep{\fill}}p{1.6cm} p{1.8cm} p{1.8cm} p{2.6cm} p{3.2cm} p{2.4cm}@{}}
\toprule
\textbf{Author(s), Year} & \textbf{Focus} & \textbf{Method} & \textbf{Variables} & \textbf{Main Findings} & \textbf{Limitations} \\
\midrule
Bellaiche et al. (2023) (N=297) & Human vs. AI art preference & Two experimental studies & Human vs. AI source cues, preference ratings & Strong bias toward human-labeled art across all criteria (liking, beauty, profundity, worth), explained by authenticity and perceived effort & All artworks were actually AI-created; used deception about origins \\
\midrule
Epstein et al. (2020) (N=433) & Authorship credit for AI art & Two experimental studies & AI anthropomorphism, responsibility attribution & Credit allocation affected by how AI is described; anthropomorphic language increases AI responsibility attribution & Pre-generative AI era; vignette-based tasks; no real artworks \\
\midrule
He et al. (2025) (N=155) & Human–AI co-creation & Online vignette experiment & Contribution types, attribution & Visible human effort increased human credit; opaque effort favored AI & Scenario-based, self-reported \\
\midrule
Horton et al. (2023) (N=2,965) & Creativity under AI disclosure & 6 online experiments & Creativity ratings, disclosure effects & AI-labeled art judged less creative; human-labeled more creative by contrast & Focus on creativity only; limited valuation \\
\midrule
Lee et al. (2024) (auction records) & Art market dynamics & Auction data analysis & Social vs. visual features & Social signals (reputation, sales history) influenced prices more than visual features & Established markets only; no AI art \\
\midrule
Lima et al. (2025) (N=432) & authorship and rights attribution & Juried art competition & Creativity, effort, quality & egocentric effect in authorship and rights attribution, & Held Online exhibition \\
\midrule
Messer (2024) (N=560) & AI involvement levels & 3 vignette experiments & Human-AI co-creation levels & More visible human input increases perceived creativity and value & Hypothetical scenarios only \\

\midrule
van Hees et al. (2025) (N=264) & AI art perception & 2 comparison experiments & Preference, detection ability & Unlabeled AI sometimes preferred; participants detect AI above chance & Visual judgments only; DALL·E 2 limited \\
\midrule
\textbf{This study (N=259)} & \textbf{Value \& reward allocation} & \textbf{In-person juried competition with parallel creator/evaluator streams} & \textbf{Creativity, effort, technical skill, visual appeal, price, rights attribution, prize allocation} & \textbf{Creators price by effort invested; evaluators price by artistic merit; training data artists marginalized in favor of creators and tech entities; legal backgrounds favor distributed rewards; CS backgrounds reduce creator compensation} & \textbf{Held in-person exhibition; primarily Art/CS/Law students from single university; hypothetical pricing} \\
\bottomrule
\end{tabular*}
\end{table*}

\begin{table*}[htbp]
\centering
\captionsetup{labelsep=colon}
\caption{Codebook sample for qualitative analysis of creator responses (N = 104). Codes were developed through iterative open coding and organized into five thematic categories.}
\label{app:codebook}
\small
\renewcommand{\arraystretch}{1.2}
\begin{tabular}{@{}p{1.05in}p{1.9in}p{3.4in}@{}}
\toprule
\textbf{Code} & \textbf{Definition} & \textbf{Sample Coding} \\
\midrule

\multicolumn{3}{@{}l}{\textit{\textbf{Process}}} \\[2pt]
Ideation & Pre-existing ideas prior to AI interaction & ``I uploaded the image and gave the rough prompt: try to make an image of the building of the line drawing uploaded, but entirely out of food'' (P14) \\[4pt]
Iteration & Repeated attempts without specifying a problem & ``I input over 100 prompts, each one as they occurred to me, and often guided the AI toward or away from decisions it seemed to have made'' (P89) \\[4pt]
AI-Driven & Output primarily determined by model & ``I tried to think of some random combinations of objects, themes, and styles to throw at the AI generator to see how it would handle them'' (P55) \\
\midrule

\multicolumn{3}{@{}l}{\textit{\textbf{Challenges}}} \\[2pt]
Technical Limitations & Model constraints or artifacts & ``The limitations of the image generation model to create images as I described forced me to pivot and adjust my language multiple times'' (P2) \\[4pt]
Prompting & Writing/refining prompts to steer output & ``AI often misunderstood my prompting, so I tried to be more clear and also give more creative freedom to the AI, instead of focusing on my own vision'' (P11) \\[4pt]
Consistency & Maintaining coherence across iterations & ``When asking for Aphrodite to be added, I got a completely different image that I had not wanted, though the goddess was added correctly'' (P41) \\[4pt]
Adaptation & Adjusting approach after constraints & ``I found that overly lengthy prompts led to inaccurate depictions and were often less desirable than a concise and well-thought-out prompt'' (P82) \\
\midrule

\multicolumn{3}{@{}l}{\textit{\textbf{Inspiration}}} \\[2pt]
Art / Art History & References to artists or movements & ``Masters of light such as Monet, Manet, and Van Gogh inspired my approach---using dynamic brushstrokes and vibrant tones to evoke movement and atmosphere'' (P1) \\[4pt]
Personal Reflection & Memories, emotions, lived experience & ``The main inspirations were based purely on how calming and charming the window art in Dahlgren Chapel was. Strawberries are a lot sweeter back home'' (P87) \\[4pt]
Academic Subjects & Disciplinary content informing concept & ``I instructed the AI to disregard previous inputs, then told it to generate an image based on a poem I had previously written for another project'' (P5) \\[4pt]
Culture & Traditions, heritage, popular culture & ``Earthbound on the Super Nintendo and Gravity Falls on Disney Channel inspired my thematic decision and art style for my submission'' (P69) \\[4pt]
Current Events & News, social issues, philosophy & ``It gave me the idea to create an artwork showing the possibilities that AI tools could unlock for humanity, ascribing hope to the future of AI'' (P20) \\
\midrule

\multicolumn{3}{@{}l}{\textit{\textbf{Value}}} \\[2pt]
Uniqueness & Claims about originality or distinctiveness & ``I believe my AI art has financial value based on its originality and its novelty as a proof of concept of AI being able to challenge established brightlines of genre in art'' (P62) \\[4pt]
Idea / Concept & Theme, narrative, or central message & ``Perhaps yes, because it evokes thought. It would steal my attention and start a conversation. In my opinion, that is what gives it value'' (P22) \\[4pt]
Market Forces & Saleability, demand, monetization & ``AI-generated art is gaining recognition in digital art markets, and demand for thought-provoking, high-quality AI pieces continues to grow'' (P68) \\[4pt]
Accessibility & Ease of reproduction; exclusivity concerns & ``Any sort of art can have financial value if you find the right niche and promote it in the right way'' (P83) \\[4pt]
Data Ethics & Consent, attribution of training data & ``The credit and compensation for its development should partly go to the artists whose work the AI copied, likely without attribution'' (P49) \\
\midrule

\multicolumn{3}{@{}l}{\textit{\textbf{Skills Utilized}}} \\[2pt]
Creative Direction & Intentional planning, aesthetic judgment & ``I decided on historical accuracy, emotional atmosphere, character attire, vintage color tones, and balanced composition to create an immersive scene'' (P37) \\[4pt]
Prompt Engineering & Parameters, seeds, negative prompts & ``I led the creative process, refining prompts and selecting the most powerful composition'' (P68) \\[4pt]
Artistic Techniques & Composition, color, post-processing & ``After generating the initial image, I color-graded, filtered, and refined the visuals to define focal points and convey a different tone'' (P66) \\[4pt]
Problem-solving & Iteration to fix specific problems & ``These later iterations from these conceptual prompts got further away from the style that I sought, and I did not submit these iterations'' (P74) \\[4pt]
Writing & Artist statements, titling, curatorial text & ``I titled it and wrote a statement to guide interpretation'' \\
\bottomrule
\end{tabular}
\end{table*}



\label{app:creativity_perceptions}
\begin{table}
\captionsetup{labelsep=colon}
\caption{Creativity perception models. We model perceived creativity separately for
creators and evaluators. Blank cells indicate covariates not included in a given specification.
Refer to Section~\ref{sec:rq1} for details. Standard errors are shown inside parentheses.
$^{*}\,p<0.05$; $^{**}\,p<0.01$; $^{***}\,p<0.001$.}
\centering
\small
\begin{tblr}[         
caption={Creativity Perceptions: Creator vs Evaluator Models},
note{}={* p < 0.05, ** p < 0.01, *** p < 0.001},
]                     
{                     
colspec={Q[]Q[]Q[]},
column{2-3}={}{halign=c,},
column{1}={}{halign=l,},
hline{22}={1-3}{solid, black, 0.05em},
}                     
\toprule
& Creators: Creativity Perceptions & Evaluators: Creativity Perceptions \\ \midrule 
(Intercept) & \num{-0.785} & \num{-0.290}* \\
& (\num{0.645}) & (\num{0.127}) \\
Effort & \num{0.206} & \num{0.432}*** \\
& (\num{0.113}) & (\num{0.047}) \\
Visual appeal &  & \num{0.254}*** \\
&  & (\num{0.041}) \\
Technical quality & \num{0.246}* & \num{0.167}*** \\
& (\num{0.094}) & (\num{0.048}) \\
AI usage frequency & \num{-0.086} & \num{0.049} \\
& (\num{0.104}) & (\num{0.059}) \\
Display Intent: Museum & \num{0.343} &  \\
& (\num{0.224}) &  \\
Prompts (word count) & \num{0.422}* &  \\
& (\num{0.186}) &  \\
Art background & \num{-0.088} & \num{-0.052} \\
& (\num{0.218}) & (\num{0.117}) \\
CS background & \num{0.391} & \num{-0.125} \\
& (\num{0.228}) & (\num{0.123}) \\
Law background & \num{0.205} & \num{0.120} \\
& (\num{0.223}) & (\num{0.130}) \\
Num.Obs. & \num{99} & \num{536} \\
R2 & \num{0.382} &  \\
R2 Adj. & \num{0.327} &  \\
R2 Marg. &  & \num{0.558} \\
R2 Cond. &  & \num{0.607} \\
AIC & \num{288.6} & \num{1638.5} \\
BIC & \num{314.6} & \num{1685.7} \\
ICC &  & \num{0.1} \\
Log.Lik. & \num{-134.313} &  \\
RMSE & \num{0.94} & \num{0.97} \\
\bottomrule
\end{tblr}
\end{table}


\label{app:price}
\begin{table}[h]
\label{tab:price}
\captionsetup{labelsep=colon}
\caption{Pricing models. We model pricing separately for creators and evaluators (evaluator mixed-effects model). Refer to Section~\ref{sec:rq1}
for details. Standard errors are shown inside parentheses. $^{*}\,p<0.05$;
$^{**}\,p<0.01$; $^{***}\,p<0.001$.}
\centering
\footnotesize
\begin{tblr}[         
caption={Evaluator vs Creator Pricing Models},
note{}={* p < 0.05, ** p < 0.01, *** p < 0.001},
]                     
{                     
colspec={Q[]Q[]Q[]},
column{2-3}={}{halign=c,},
column{1}={}{halign=l,},
hline{30}={1-3}{solid, black, 0.05em},
}                     
\toprule
& Creators: Pricing & Evaluators: Pricing \\ \midrule 
(Intercept) & \num{-0.605} & \num{0.468}*** \\
& (\num{0.868}) & (\num{0.138}) \\
Creativity & \num{-0.162} & \num{0.137}*** \\
& (\num{0.145}) & (\num{0.030}) \\
Effort & \num{0.133} & \num{0.127}*** \\
& (\num{0.156}) & (\num{0.035}) \\
Visual appeal &  & \num{0.191}*** \\
&  & (\num{0.029}) \\
Technical quality & \num{-0.003} & \num{0.054} \\
& (\num{0.126}) & (\num{0.033}) \\
Art background & \num{-0.577}* & \num{-0.091} \\
& (\num{0.281}) & (\num{0.134}) \\
CS background & \num{0.128} & \num{-0.381}** \\
& (\num{0.301}) & (\num{0.141}) \\
Law background & \num{0.192} & \num{0.043} \\
& (\num{0.297}) & (\num{0.150}) \\
Prompts (word count) & \num{0.643}** &  \\
& (\num{0.236}) &  \\
AI usage frequency & \num{-0.092} & \num{0.083} \\
& (\num{0.142}) & (\num{0.068}) \\
Display Intent: Commercial & \num{0.048} &  \\
& (\num{0.283}) &  \\
Display Intent: Gallery & \num{0.391} &  \\
& (\num{0.368}) &  \\
Display Intent: Museum & \num{0.354} &  \\
& (\num{0.396}) &  \\
Display Intent: Public & \num{-0.058} &  \\
& (\num{0.327}) &  \\
Num.Obs. & \num{85} & \num{535} \\
R2 & \num{0.280} &  \\
R2 Adj. & \num{0.160} &  \\
R2 Marg. &  & \num{0.446} \\
R2 Cond. &  & \num{0.693} \\
AIC & \num{277.9} & \num{1278.9} \\
BIC & \num{312.1} & \num{1330.3} \\
ICC &  & \num{0.4} \\
Log.Lik. & \num{-124.952} &  \\
RMSE & \num{1.05} & \num{0.60} \\
\bottomrule
\end{tblr}
\end{table}

\label{app:prize}
\begin{table}[h]
\captionsetup{labelsep=colon} 
\caption{Prize allocation models. We estimate prize allocation for User, Artist, and
Tech entities for: (C) creator and (E)
evaluator models. Refer to Section~\ref{sec:rq2} for details. Standard errors are
shown inside parentheses. $+\,p<0.1$; $^{*}\,p<0.05$; $^{**}\,p<0.01$; $^{***}\,p<0.001$.}
\centering
\small
\begin{tblr}[         
note{}={+ p \num{< 0.1}, * p \num{< 0.05}, ** p \num{< 0.01}, *** p \num{< 0.001}},
]                     
{                     
colspec={Q[]Q[]Q[]Q[]Q[]Q[]Q[]},
column{2-7}={}{halign=c,},
column{1}={}{halign=l,},
hline{28}={1-7}{solid, black, 0.05em},
}                     
\toprule
& User (C) & User (E) & Artist (C) & Artist (E) & Tech (C) & Tech (E) \\ \midrule 
Creativity & \num{0.020} & \num{0.185} & \num{0.013} & \num{-0.041} & \num{0.003} & \num{0.096} \\
& (\num{0.023}) & (\num{0.141}) & (\num{0.041}) & (\num{0.150}) & (\num{0.045}) & (\num{0.154}) \\
Effort & \num{0.015} & \num{0.256} & \num{-0.004} & \num{-0.213} & \num{0.005} & \num{-0.527}** \\
& (\num{0.025}) & (\num{0.167}) & (\num{0.045}) & (\num{0.179}) & (\num{0.049}) & (\num{0.186}) \\
Technical quality & \num{0.034} & \num{-0.221} & \num{-0.036} & \num{0.236} & \num{0.011} & \num{0.493}** \\
& (\num{0.021}) & (\num{0.159}) & (\num{0.038}) & (\num{0.170}) & (\num{0.042}) & (\num{0.179}) \\
Visual appeal &  & \num{0.293}* &  & \num{0.263}+ &  & \num{-0.091} \\
&  & (\num{0.143}) &  & (\num{0.156}) &  & (\num{0.154}) \\
Art background & \num{-0.010} & \num{-0.622} & \num{0.075} & \num{1.293} & \num{-0.053} & \num{0.130} \\
& (\num{0.048}) & (\num{0.611}) & (\num{0.084}) & (\num{0.942}) & (\num{0.093}) & (\num{0.912}) \\
CS background & \num{-0.003} & \num{-1.711}** & \num{0.016} & \num{-0.881} & \num{0.005} & \num{-1.515} \\
& (\num{0.051}) & (\num{0.651}) & (\num{0.090}) & (\num{0.977}) & (\num{0.099}) & (\num{0.957}) \\
Law background & \num{0.040} & \num{0.138} & \num{0.131} & \num{2.217}* & \num{-0.009} & \num{1.045} \\
& (\num{0.050}) & (\num{0.678}) & (\num{0.088}) & (\num{1.062}) & (\num{0.098}) & (\num{1.039}) \\
AI usage frequency & \num{0.040}+ & \num{0.099} & \num{-0.073}+ & \num{0.016} & \num{-0.101}* & \num{-0.177} \\
& (\num{0.024}) & (\num{0.297}) & (\num{0.042}) & (\num{0.466}) & (\num{0.046}) & (\num{0.455}) \\
Total Prompts(wordcount) & \num{-0.006} &  & \num{0.015} &  & \num{-0.118} &  \\
& (\num{0.042}) &  & (\num{0.074}) &  & (\num{0.081}) &  \\
Display Intent: Museum & \num{-0.038} & \num{0.453}* & \num{0.003} & \num{-0.384} & \num{0.015} & \num{0.205} \\
& (\num{0.070}) & (\num{0.180}) & (\num{0.123}) & (\num{0.269}) & (\num{0.136}) & (\num{0.265}) \\
Display Intent: Commercial & \num{-0.053} &  & \num{0.225}** &  & \num{0.030} &  \\
& (\num{0.049}) &  & (\num{0.087}) &  & (\num{0.096}) &  \\
Display Intent: Gallery & \num{0.055} &  & \num{-0.062} &  & \num{-0.075} &  \\
& (\num{0.064}) &  & (\num{0.112}) &  & (\num{0.124}) &  \\
Display Intent: Public & \num{0.108}+ &  & \num{0.029} &  & \num{0.016} &  \\
& (\num{0.056}) &  & (\num{0.098}) &  & (\num{0.108}) &  \\
Num.Obs. & \num{99} & \num{536} & \num{99} & \num{536} & \num{99} & \num{536} \\
R2 & \num{0.168} &  & \num{0.135} &  & \num{0.097} &  \\
R2 Marg. &  & \num{0.205} &  & \num{0.121} &  & \num{0.067} \\
R2 Cond. &  & \num{0.694} &  & \num{0.831} &  & \num{0.812} \\
AIC & \num{-9.9} & \num{466.7} & \num{101.5} & \num{479.2} & \num{121.7} & \num{431.4} \\
BIC & \num{26.4} & \num{513.8} & \num{137.8} & \num{526.3} & \num{158.0} & \num{478.5} \\
ICC &  & \num{0.6} &  & \num{0.8} &  & \num{0.8} \\
Log.Lik. & \num{18.959} &  & \num{-36.747} &  & \num{-46.851} &  \\
RMSE & \num{0.20} & \num{0.29} & \num{0.35} & \num{0.26} & \num{0.39} & \num{0.25} \\
\bottomrule
\end{tblr}
\label{tab:prize}
\end{table}

\clearpage
\label{app:rights}
\begin{table}[htbp]
\captionsetup{labelsep=colon} 
\caption{Exclusive Rights Attribution Models. Attribution of exclusive rights
for Users, Tech entities (Tech), and training-data Artist from creators and evaluators. Standard errors are shown inside parentheses. $+\,p<0.1$; $^{*}\,p<0.05$;
$^{**}\,p<0.01$; $^{***}\,p<0.001$.}
\centering
\small
\begin{tblr}[         
note{}={+ p \num{< 0.1}, * p \num{< 0.05}, ** p \num{< 0.01}, *** p \num{< 0.001}},
note{ }={C = Creators, E = Evaluators},
note{  }={Tech = AI companies, developers, and model creators},
note{   }={Artist = Training data artists},
]                     
{                     
colspec={Q[]Q[]Q[]Q[]Q[]Q[]Q[]},
column{2-7}={}{halign=c,},
column{1}={}{halign=l,},
hline{26}={1-7}{solid, black, 0.05em},
}                     
\toprule
& User (C) & User (E) & Tech (C) & Tech (E) & Artist (C) & Artist (E) \\ \midrule 
Creativity & \num{0.48}+ & \num{0.10} & \num{0.47} & \num{0.24} & \num{0.78} & \num{0.20} \\
& (\num{0.26}) & (\num{0.15}) & (\num{0.29}) & (\num{0.25}) & (\num{0.66}) & (\num{0.24}) \\
Effort & \num{0.13} & \num{0.66}*** & \num{0.13} & \num{-0.17} & \num{-0.85} & \num{0.53}+ \\
& (\num{0.29}) & (\num{0.17}) & (\num{0.32}) & (\num{0.27}) & (\num{0.55}) & (\num{0.28}) \\
Visual appeal &  & \num{0.35}* &  & \num{-0.32} &  & \num{0.34} \\
&  & (\num{0.15}) &  & (\num{0.24}) &  & (\num{0.24}) \\
Technical quality & \num{0.16} & \num{-0.27} & \num{-0.08} & \num{0.60}* & \num{-0.47} & \num{-0.53}+ \\
& (\num{0.24}) & (\num{0.17}) & (\num{0.26}) & (\num{0.25}) & (\num{0.47}) & (\num{0.29}) \\
Art background & \num{0.54} & \num{-0.72} & \num{-0.17} & \num{0.10} & \num{-0.82} & \num{-0.82} \\
& (\num{0.58}) & (\num{0.70}) & (\num{0.62}) & (\num{1.29}) & (\num{1.18}) & (\num{1.47}) \\
CS background & \num{1.26}* & \num{-0.22} & \num{-0.33} & \num{-0.62} & \num{-0.36} & \num{-1.62} \\
& (\num{0.64}) & (\num{0.74}) & (\num{0.62}) & (\num{1.40}) & (\num{1.08}) & (\num{1.57}) \\
Law background & \num{-0.10} & \num{1.42}+ & \num{-0.83} & \num{1.21} & \num{0.07} & \num{5.17}* \\
& (\num{0.60}) & (\num{0.78}) & (\num{0.68}) & (\num{1.37}) & (\num{1.08}) & (\num{2.13}) \\
AI usage frequency & \num{-0.13} & \num{0.22} & \num{0.13} & \num{-0.21} & \num{-0.14} & \num{-0.10} \\
& (\num{0.32}) & (\num{0.34}) & (\num{0.28}) & (\num{0.66}) & (\num{0.46}) & (\num{0.73}) \\
Display Intent: Museum & \num{-0.65} & \num{0.61}** & \num{-0.13} & \num{0.19} & \num{-1.33} & \num{0.16} \\
& (\num{0.89}) & (\num{0.22}) & (\num{0.95}) & (\num{0.41}) & (\num{1.62}) & (\num{0.42}) \\
Display Intent: Commercial & \num{-0.82} &  & \num{-0.54} &  & \num{1.47} &  \\
& (\num{0.65}) &  & (\num{0.60}) &  & (\num{1.20}) &  \\
Display Intent: Gallery & \num{0.56} &  & \num{-1.12} &  & \num{0.14} &  \\
& (\num{0.79}) &  & (\num{0.84}) &  & (\num{1.40}) &  \\
Display Intent: Public & \num{1.07}+ &  & \num{-0.43} &  & \num{1.52} &  \\
& (\num{0.64}) &  & (\num{0.66}) &  & (\num{1.27}) &  \\
Num.Obs. & \num{99} & \num{536} & \num{99} & \num{536} & \num{99} & \num{536} \\
R2 Marg. &  & \num{0.287} &  & \num{0.017} &  & \num{0.121} \\
R2 Cond. &  & \num{0.791} &  & \num{0.944} &  & \num{0.952} \\
AIC & \num{122.2} & \num{504.5} & \num{118.4} & \num{251.8} & \num{69.9} & \num{309.1} \\
BIC & \num{158.5} & \num{551.6} & \num{154.8} & \num{299.0} & \num{106.2} & \num{356.2} \\
ICC &  & \num{0.7} &  & \num{0.9} &  & \num{0.9} \\
Log.Lik. & \num{-47.091} &  &  &  & \num{-20.927} &  \\
RMSE & \num{0.40} & \num{0.27} & \num{0.38} & \num{0.19} & \num{0.26} & \num{0.19} \\
\bottomrule
\end{tblr}
\end{table}


\end{document}